\documentclass{article}

\usepackage{microtype}
\usepackage{graphicx}
\usepackage[justification=centering,singlelinecheck=false]{subfig}
\usepackage{caption}

\usepackage{booktabs} 

\usepackage{hyperref}

\usepackage{amsfonts}       
\usepackage[nice]{nicefrac}       
\usepackage{xcolor}         
\usepackage{csquotes}
\usepackage{enumitem}
\usepackage{amssymb}
\usepackage{caption}
\usepackage{color}
\usepackage{tikz}
\usepackage{verbatim}
\usepackage{pgf}
\usepackage{pgfplots}
\usepackage{lmodern}

\usepackage{amsmath}
\usepackage{mathtools}

\usetikzlibrary{positioning}

\usepackage{pgfplots}
\pgfplotsset{width=10cm,compat=1.9}

\usepgfplotslibrary{external}
\usepackage[accepted]{icml2023}

\newcommand{\E}{\mathbb{E}}
\newcommand{\Eb}[1]{\E\big(#1\big)}
\newcommand{\bb}[1]{\big(#1\big)}
\newcommand{\Prob}{\operatorname{P}}

\newcommand{\D}{\mathcal{D}}

\newcommand{\R}{\mathbb{R}}

\newcommand{\var}{\mathrm{Var}}

\newcommand{\x}{\mathbf{x}}
\newcommand{\abf}{\mathbf{a}}
\newcommand{\w}{\mathbf{w}}

\newcommand{\indep}{\perp \!\!\! \perp}
\newcommand{\dep}{\not \! \indep}

\icmltitlerunning{Theoretical Behavior of XAI Methods in the Presence of Suppressor Variables}

\begin{document}

\twocolumn[
\icmltitle{Theoretical Behavior of XAI Methods in the Presence of Suppressor Variables}



\icmlsetsymbol{equal}{*}

\begin{icmlauthorlist}
\icmlauthor{Rick Wilming}{tub}
\icmlauthor{Leo Kieslich}{tub}
\icmlauthor{Benedict Clark}{ptb}
\icmlauthor{Stefan Haufe}{tub,ptb,charite}
\end{icmlauthorlist}

\icmlaffiliation{tub}{Technische Universität, Berlin, Germany}
\icmlaffiliation{ptb}{Physikalisch-Technische Bundesanstalt, Berlin, Germany}
\icmlaffiliation{charite}{Charité – Universitätsmedizin, Berlin, Germany}

\icmlcorrespondingauthor{Stefan Haufe}{haufe@tu-berlin.de}

\icmlkeywords{XAI, interpretability, suppressor variable, theory}

\vskip 0.3in
]



\printAffiliationsAndNotice{} 

\begin{abstract}
In recent years, the community of `explainable artificial intelligence' (XAI) has created a vast body of methods to bridge a perceived gap between model `complexity' and `interpretability'.
However, a concrete problem to be solved by XAI methods has not yet been formally stated. As a result, XAI methods are lacking theoretical and empirical evidence for the `correctness' of their explanations, limiting their potential use for quality-control and transparency purposes.
At the same time, \citet{haufeInterpretationWeightVectors2014} showed, using simple toy examples, that even standard interpretations of linear models can be highly misleading. Specifically, high importance may be attributed to so-called suppressor variables lacking any statistical relation to the prediction target. This behavior has been confirmed empirically for a large array of XAI methods in \citet{wilmingScrutinizingXAIUsing2022}. Here, we go one step further by deriving analytical expressions for the behavior of a variety of popular XAI methods on a simple two-dimensional binary classification problem involving Gaussian class-conditional distributions. We show that the majority of the studied approaches will attribute non-zero importance to a non-class-related suppressor feature in the presence of correlated noise. This poses important limitations on the interpretations and conclusions that the outputs of these XAI methods can afford.
\end{abstract}

\section{Introduction}\label{sec:introduction}
The field of `explainable artificial intelligence' (XAI) is devoted to answering the broad question of why
an automatic decision system put forward a certain prediction. This is often addressed by techniques that attribute a so-called `importance' score to each feature of an individual test input.
It is commonly agreed that being able to answer this question is necessary to create trust in and a better understanding of the behavior of such decision systems~\cite{baehrensHowExplainIndividual2010,ribeiroWhyShouldTrust2016,binderLayerWiseRelevancePropagation2016,lundbergUnifiedApproachInterpreting2017,fisherAllModelsAre2019}.
In \citet{haufeInterpretationWeightVectors2014} and \citet{wilmingScrutinizingXAIUsing2022}, it was shown that features which certain XAI methods determine to be important, e.g. by inspecting their corresponding weights of a linear model, may actually not have any statistical association with the predicted variable. As a result, the provided `explanation' may not agree with prior domain knowledge of an expert user and might undermine that user's trust in the predictive model, even if it performs optimally. Indeed, a highly accurate model might exploit so-called suppressor features~\cite{congerRevisedDefinitionSuppressor1974, friedmanGraphicalViewsSuppression2005}, which can be statistically independent of the prediction target yet still lead to increased prediction performance. On the other hand, incorrect explanations may implant misconceptions about the data, the model and/or the relationship between the two into a user's mind, which could lead to misguided actions that could be harmful.

While \citet{haufeInterpretationWeightVectors2014} have introduced low-dimensional and well-controlled examples to illustrate the problem of suppressor variables for model interpretation, \citet{wilmingScrutinizingXAIUsing2022} showed empirically that the emergence of suppressors indeed poses a problem for a large group of XAI methods and diminishes their `explanation performance'. Here, we go one step further and derive analytical expressions for 
commonly used XAI methods for a simple two-dimensional linear data generation process capable of creating suppressor variables by parametrically inducing correlations between features. In particular, we investigate which XAI approaches attribute non-zero importance to plain suppressor variables that are by construction independent of the prediction target and thereby violate a data-driven definition of feature importance recently put forward by \citet{wilmingScrutinizingXAIUsing2022}.


\section{Related Work}\label{sec:related-work}
XAI methods often analyze ML models in a post-hoc manner~\cite{arrietaExplainableArtificialIntelligence2019}, where a trained model deemed to be `non-interpretable', such as a deep neural network, is given, while the XAI methods attempt to `reverse-engineer' its decision for a given input sample. A crucial limitation of the field of XAI is that it is still an open question what formal requirements \textit{correct} explanations would need to fulfill and what conclusions about data, model, and their relationship the analysis of an importance map provided by XAI methods should afford.
The lack of a clear definition of what problem XAI is supposed to solve led to multiple studies evaluating explanation methods~\citep[e.g.][]{doshi-velezRigorousScienceInterpretable2017, kimInterpretabilityFeatureAttribution2018, alvarez-melisRobustnessInterpretabilityMethods2018, adebayoSanityChecksSaliency2018, sixtWhenExplanationsLie2020}.
Yet, these studies primarily employ auxiliary metrics to measure secondary quality aspects, such as the stability of the provided maps. For example, \citet{yang2019benchmarking} investigate how importance maps for one model change relative to another model.  Until recently, it has been considered difficult to define and evaluate the correctness of importance maps, because real-world datasets, which are ubiquitous in the ML community as benchmarks for supervised prediction tasks, do not offer access to the `true' set of important features. However, several XAI benchmarks using controlled synthetic data have emerged in the past three years.
\citet{agarwal2022openxai} propose a benchmark that can generate synthetic data and assess XAI methods on a broad set of evaluation metrics. The authors state that their framework predominantly serves the purpose of gaining a better understanding of a model's internal mechanics, which would primarily show the debugging capabilities of XAI methods rather than their ability to generate knowledge of `real-world' effects.
\citet{sixtWhenExplanationsLie2020} provide a theoretical analysis of convergence problems of so-called saliency methods, especially Layer-wise Relevance Propagation
\citep[LRP,][]{bachPixelWiseExplanationsNonLinear2015}, Deep Taylor
Decomposition \citep[DTD,][]{montavonExplainingNonLinearClassification2017}, and DeepLIFT \citep{shrikumarLearningImportantFeatures2017}. Notably, the provided derivations do not take the model's input data into account.
\citet{kindermansLearningHowExplain2017} use a minimal data generation example, to mainly motivate a discussion about drawbacks of saliency maps to finally propose novel explanation techniques based on the DTD framework.
\citet{janzingFeatureRelevanceQuantification2020} consider a structural data generation model, promoting unconditional expectations as a value function for SHAP~\citep{lundbergUnifiedApproachInterpreting2017} by demonstrating that observational conditional expectations are flawed.
In an extensive study on Partial Dependency Plots~\cite{friedmanGreedyFunctionApproximation2001} and M-plots~\cite{apleyVisualizingEffectsPredictor2019}, \citet{grompingModelagnosticEffectsPlots2020} theoretically analyse a regression task via a pre-defined regression model $\E(Y|\x)$ with multivariate Gaussian distributed data. They argue that M-plots can lead to deceptive results, especially if machine learning models rely on interaction effects. 
\citet{wilmingScrutinizingXAIUsing2022} empirically study common post-hoc explanation methods using a carefully crafted dataset based on a linear data generation process. Here, all statistical dependencies and absolute feature importances are well defined, giving rise to ground-truth 
importance maps.
This empirical study showed that most XAI methods indeed highlight suppressor features as important.

\subsection{Definition of Feature Importance}\label{sec:feature-importance}
In this paper, we adopt a data-driven notion proposed by \citet{wilmingScrutinizingXAIUsing2022} as a tentative definition of feature importance. We
consider a supervised learning task, where a model $f: \R^d \to \R$ learns a function between an input $\x^{(i)} \in \R^d$ and a target $y^{(i)} \in \R$, based on training data $\D = \{(\x^{(i)}, y^{(i)}\}_{i=1}^{N}$. Here, $\x^{(i)}$ and $y^{(i)}$ are realizations of the random variables $\mathbf{X}$ and $Y$, with joint probability density function $p_{\mathbf{X},Y}(\x,y)$.
Then a feature $X_j$ can be defined to be important if it has a  statistical association to the target variable $Y$, i.e.
\begin{align}
X_j \text{ is important} \Rightarrow X_j \dep Y.
\label{eq:importance_definition}
\end{align}


\subsection{Suppressor Variables}\label{subsec:suppressor-variables}
\begin{figure}
    \centering
    \subfloat[Confounder]{\label{fig:confounder-collider-a}\includegraphics[scale=0.225]{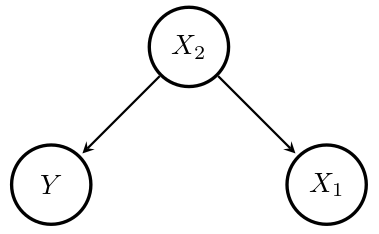}}
\hspace{0.05\columnwidth}
    \subfloat[Suppressor and Collider]{\label{fig:confounder-collider-b}\includegraphics[scale=0.225]{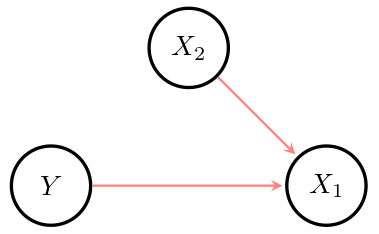}}
    \caption{
    In (a), feature $X_2$ is a confounder variable influencing $Y$ and another feature $X_1$, causing spurious associations. In contrast, in (b) $X_2$ is a so-called suppressor variable that has no statistical association with the target $Y$, although both influence feature $X_1$, which is called a collider.}
    \label{fig:confounder-collider}
\end{figure}
To illustrate the characteristics of suppressor variables, consider a binary classification problem with two measured scalar input features $x_1$ and $x_2$, where $x_1$ carries all discriminative information, following \citet{haufeInterpretationWeightVectors2014}. We design the input data such that $x_1$ holds the signal of interest $z \in \{-1, 1\}$, which is identical to the target variable $y = z$. 
Furthermore, during the measuring process, feature $x_1$ is inadvertently obfuscated by a \textit{distractor} $\eta$: $x_1 = z+\eta$.
The second feature only consists of the distractor signal, i.e. $x_2 = \eta$.
Our goal is to learn a function that can discriminate between the two states $y=-1$ and $y=1$ or, in other words, recover the signal of interest $z$.
We can build a model solely based on feature $x_1$ to solve the classification problem, as $x_1$ is the only feature that contains information about $y=z$.
Yet, the obfuscation of $x_1$ by the distractor $\eta$ diminishes its predictive power.
On the other hand, feature $x_2$ does not contain any information about $y=z$. Therefore, a model solely based on $x_2$ cannot reach above chance-level classification accuracy.
However, a bivariate linear model with a weight vector $w=(1, -1)^{\top}$ can perfectly recover the signal of interest and, thereby, the target:
$w^\top \x = z + \eta - \eta = z = y \;$.
%
%
Additionally, Structural equation models (SEM) are depicting different ways in which a variable $X_2$ can influence the prediction of a target variable $Y$.
In Figure \ref{fig:confounder-collider-a} $X_2$ is a confounder variable influencing $Y$ and another feature $X_1$, causing spurious associations. 
Confounders can appear, for example, as watermarks in image classification tasks, as studied by \citet{lapuschkinUnmaskingCleverHans2019} and can reduce the generalization capabilities of a model to new data where confounders might be absent.
However, in contrast, we consider suppressor variables $X_2$ (see Figure \ref{fig:confounder-collider-b}) that have no statistical associations with a target variable $Y$, while $X_1$ is a collider variable, taking input from both $Y$ and $X_2$.
Here, we can establish the relation $\Prob(X_2 \mid X_1) \neq \Prob(X_2 \mid X_1, Y)$ showing a conditional dependency of the suppressor $X_2$ on the target $Y$.
These conditional dependencies are used by multivariate methods to improve the accuracy of predictions.
In practice, XAI methods do not distinguish whether a feature is a confounder or a suppressor, which can lead to misunderstandings about a model's performance and interpretation.

\section{Methods}\label{sec:methods}
The purpose of this paper is to use a simple model of suppressor variables as a device to analyze the importances produced by a number of popular XAI methods, and to compare these importance scores to our data-driven definition of feature importance \eqref{eq:importance_definition}. In the following, we introduce notation that we will use throughout the text, define the data generation model, derive the Bayes optimal classifier, and provide further technical remarks.

\subsection{Linear Generative Model} \label{subsec:linear-generative-model}
\begin{figure}
\centering
\subfloat[$c=0.8$]{
    \includegraphics[scale=0.185]{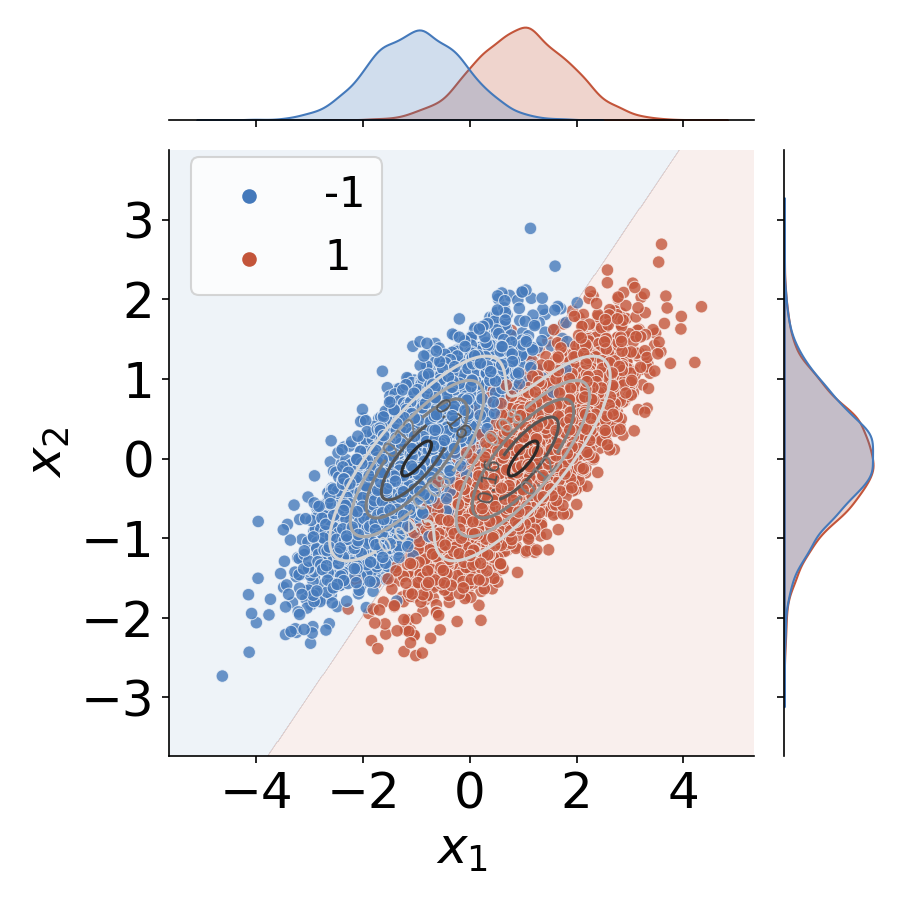}}
\subfloat[$c=0$]{
    \includegraphics[scale=0.185]{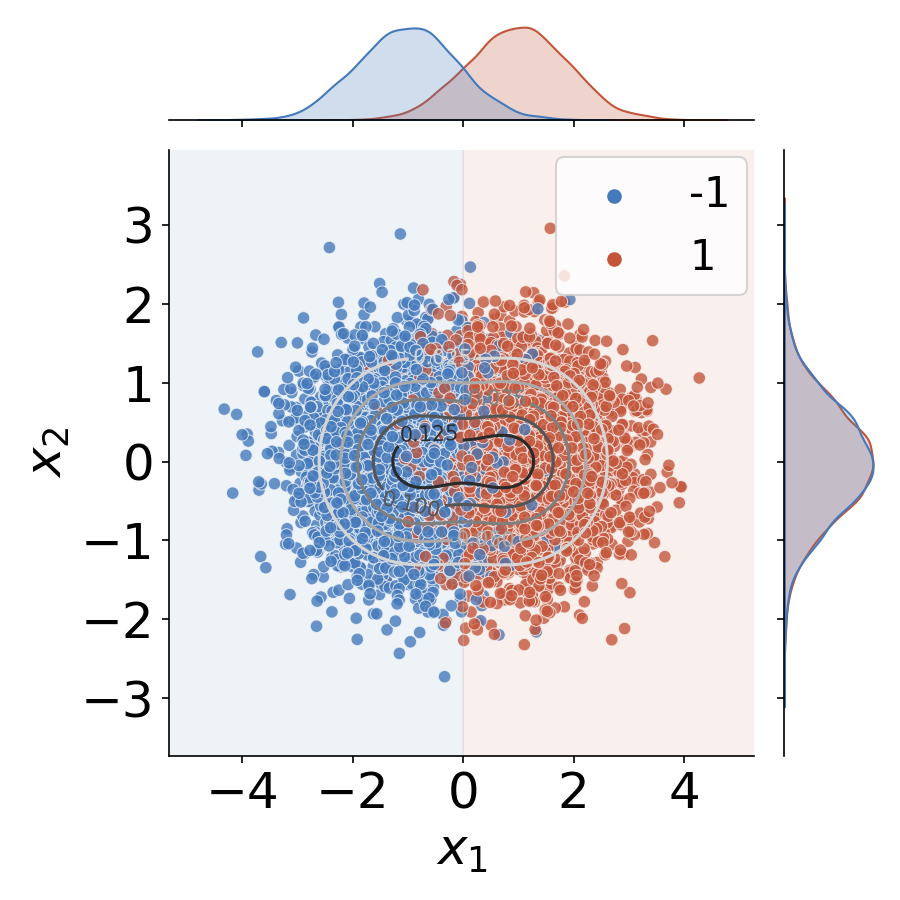}}
\subfloat[$c=-0.8$]{
    \includegraphics[scale=0.185]{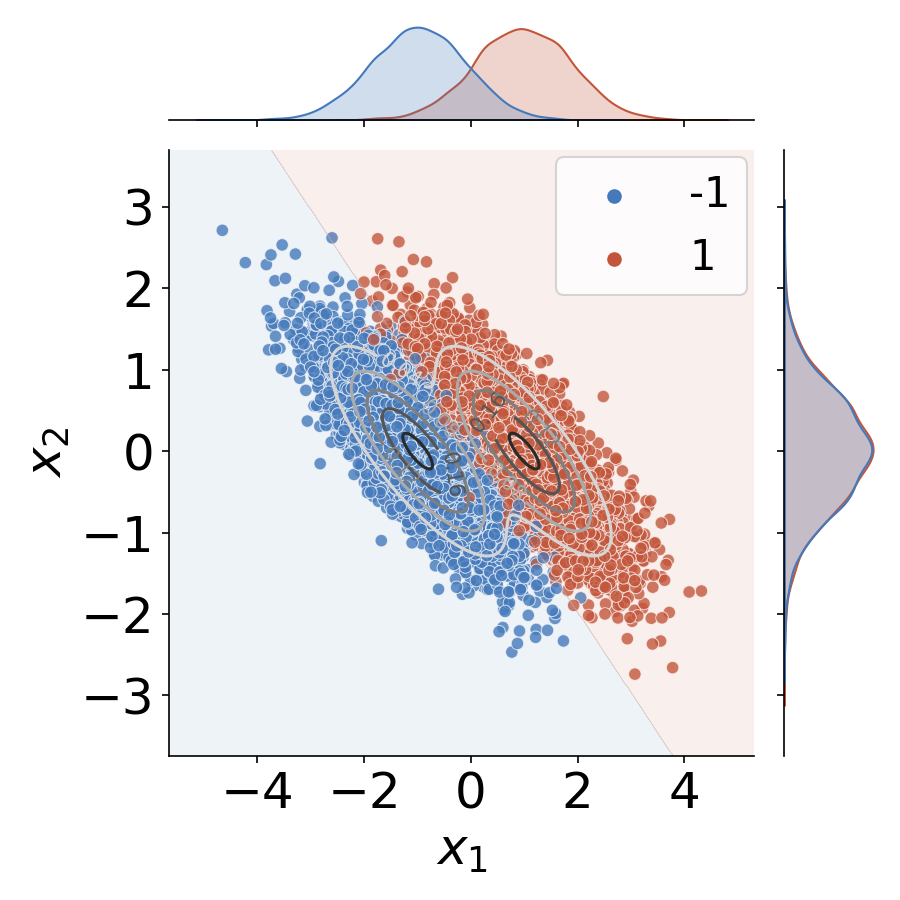}}
\caption{Data sampled from the generative process  \eqref{eq:suppressor-problem} for different correlations $c$ and constant variances $s_1^2 = 0.8$ and $s_2^2=0.5$. Boundaries of Bayes optimal decisions are shown as well. The marginal sample distributions illustrate that feature $x_2$ does not carry any class-related information.
}
\label{fig:generative_model}
\end{figure}

We now slightly extend the generative data model of the former section \ref{subsec:suppressor-variables} and provide a full specification of it. Again, we consider a binary classification problem with a two-dimensional feature space where feature $x_1$, by construction, is statistically associated with the target $y$, while feature $x_2$ fulfills the definition of a suppressor variable. Correlations between both features are introduced parametrically through a Gaussian noise process, as a result of which the Bayes optimal classifier generally needs to make use of the suppressor variable. We define $H$ and $Z$ as the random variables of the realizations $\eta$ and $z$, respectively, to describe the linear generative model
\begin{equation}
    \label{eq:suppressor-problem}
    \begin{split}
        \x = \abf z + \eta, \quad y = z,
    \end{split}
\end{equation}
with $Z \sim Rademacher(1/2)$, $\abf=(1, 0)^{\top}$ and $H \sim N(\mathbf{0}, \Sigma)$
with a covariance matrix parameterized as follows:
\begin{equation}
    \label{eq:noise-covariance}
    \Sigma = \begin{bmatrix}
                 s_1^2 & c s_1 s_2 \\ c s_1 s_2 & s_2^2 \;,
    \end{bmatrix} \;,
\end{equation}
where $s_1$ and $s_2$ are non-negative standard deviations and $c \in [-1, 1]$ is a correlation.
The vector $\abf$ is also called signal \textit{pattern}~\cite{haufeInterpretationWeightVectors2014, kindermansLearningHowExplain2017}.
With that, the generative model \eqref{eq:suppressor-problem} induces a binary classification problem, where $\mathbf{X} = (X_1, X_2)$ is the random variable of the realization $\x$ with the joint density
\begin{equation}
    \label{eq:joint-density-x1x2}
    p(\x) = \pi p_1(\x \mid Y=1) + (1-\pi) p_2(\x \mid Y=-1) \;,
\end{equation}
and prior probabilities, $\pi=\Prob(Y = \pm1) = \nicefrac{1}{2}$. The densities $p_{1/2}$ are the class-conditional densities which are both multivariate normal, with $\mathbf{X} \mid Y=y \sim N(\mu_i, \Sigma)$ for $y \in \{-1, 1\}$ and $i=1, 2$ and have identical covariance matrix $\Sigma \in \R^{2 \times 2}$ and expectations $\mu_1 = (1, 0)^{\top}$ and $\mu_2 = (-1, 0)^{\top}$. A graphical depiction of the data generated by our data model is provided in Figure~\ref{fig:generative_model}.


\subsection{Bayes Optimal Classifier}\label{subsec:bayes_optimal}
The classifier $g:\R^d \to \{-1, 1\}$ that minimizes the error $\Prob(g(\mathbf{X}) \neq Y)$ is called the Bayes optimal classifier and defined by $g(\x) = \mathbb{I}_{f^*(\x) > 1/2}$, with the conditional probability $f^*(\x)=\Prob(Y=1 | \mathbf{X}=\x)$.
For multivariate normal class-conditional densities, we can calculate the exact Bayes rule $f: \R^d \to \R$, which in this case is a linear discriminant function with $g(\x) = \mathbb{I}_{f(\x) > 0}$ and $f(\x) = \w^{\top}\x + b.$

The generative data model, defined above in section \ref{subsec:linear-generative-model}, induces a binary classification problem yielding two class-conditional densities which are both multivariate normal.
We solve the classification task in a Bayes optimal way if we assign $\x$ either to class $Y=1$ or to class $Y=-1$ based on the minimal squared Mahalanobis distance $\delta^2(\x, \mu_i) = (\x-\mu_i)^{\top} \Sigma^{-1} (\x-\mu_i)$ between $\x$ and the two class means $\mu_i, i=1,2$. 
Then the concrete form of the linear Bayes rule is determined by the coefficients
\begin{equation}
    \label{eq:bayes-optimal-weights}
    \begin{split}
        w_1 &= \alpha, \quad w_2 = -\alpha c s_1/s_2 \\
    \end{split}
\end{equation}
for $\alpha \coloneqq (1+(cs_1/s_2)^2)^{-\frac{1}{2}}$ and $||\w||_2=1$.
Note, the classification problem is set up such that the linear decision rule requires no offset or bias term, i.e. $b=0$.
In Appendix \ref{app:bayes-classifier} we provide further details for deriving the Bayes optimal decision rule $f$. 
\subsection{Notation}
Throughout, $f:\R^d \to \R$ is a learned function, in our case the Bayes optimal classifier, where $f$ usually represents the linear decision rule itself.
The dimension of the input domain, $d \in \mathbb{N}$, is set to $d=2$.
We define an index set of all features $[d] \coloneqq \{1, \dots, d\}$, in order to define features of interest as a subset $S \subset [d]$, where $x_S$ denotes the restriction of $\x \in \R^d$ to the index set $S$.
Analogously, we define the complement $C = [d] \setminus S$, defining $x_C$ as all other features that are not of interest in a particular explanation task.
We also define the output of any XAI method as a mapping $e_{S}: \R^d \to \R$ representing the importance or `relevance' assigned by the method to the feature set $S$. 

\section{Analysis of Common Explanation Methods}\label{sec:analysis}
In the following, we provide a theoretical analysis of popular XAI methods. The linear generative model \eqref{eq:suppressor-problem} is our device to assess those methods' behavior in the presence of suppressor features.


\subsection{Gradient}\label{subsec:gradient}
A ML model's gradient itself is often used for explanations, as it describes the change of the model output as a function of the change of the input parameters
\citep[e.g.][]{gevrey2003reviewVariableContr, selvarajuGradCAMVisualExplanations2017}.
For linear models, the gradient is identical to the model weights, and thus independent of the input sample. This might be in part a reason why linear models are sometimes described as `glass-box' models, particularly when it comes to explaining complex non-linear models via linear surrogate models~\citep[e.g.][]{ribeiroWhyShouldTrust2016}.
However, we can see that the Bayes optimal classifier's weights \eqref{eq:bayes-optimal-weights}, which are the gradient of the optimal decision function $f$, clearly attribute non-zero importance to the suppressor variable $x_2$, which is inconsistent with the data-driven definition of feature importance \eqref{eq:importance_definition}.
\subsection{Pattern}
\citet{haufeInterpretationWeightVectors2014} argue that the coefficients of linear models are difficult to interpret. In particular, they may highlight suppressor variables. Instead, the authors propose a transformation to convert weight vectors into parameters ${\abf}$ of a corresponding linear \textit{forward model} $\x = \abf f(\x) + \varepsilon$. The solution is provided by the covariance between the model output and each input feature: $a_j = \operatorname{Cov}(x_j, f(\x)) =  \operatorname{Cov}(x_j, f(\x)) = \operatorname{Cov} (x_j, w^{\top} \x)$, for $j=1, \dots d$, which yields a global importance map 
\begin{equation}
    \label{eq:pattern}
    e_{S}(\x) \coloneqq  (\operatorname{Cov} (\x, \x) w)_S
\end{equation}
called \textit{linear activation pattern}~\cite{haufeInterpretationWeightVectors2014}. For the generative model \eqref{eq:suppressor-problem} and the Bayes optimal classifier \eqref{eq:bayes-optimal-weights}, we obtain
\begin{equation}
    \label{eq:pattern-results}
    \begin{split}
            e_{\{1\}}(\x) = \alpha s_1^2 (1-c^2) \;, \quad
            e_{\{2\}}(\x) = 0 \;.
    \end{split}
\end{equation}
Thus, the pattern approach does not attribute any importance to the suppressor feature $x_2$.

\subsection{Faithfulness and Pixel Flipping}\label{subsec:faithfulness}
\begin{figure}
    \centering
    \subfloat[Faithfulness through\\ pixel flipping]{\label{fig:faithfulness}    \includegraphics[scale=0.225]{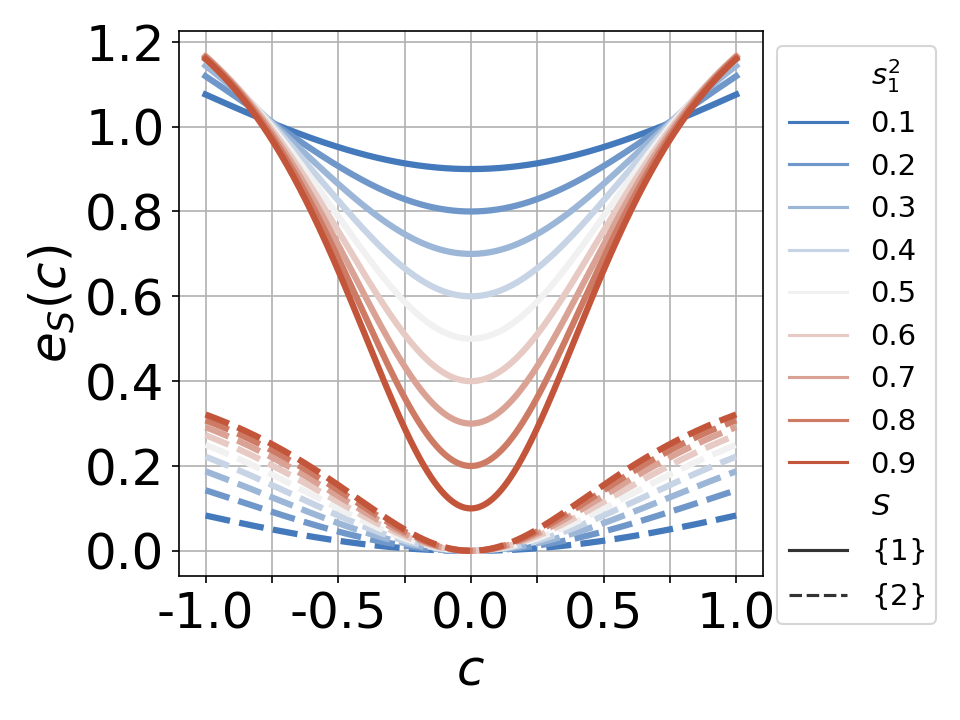}}
\hspace{0.05\columnwidth}
    \subfloat[Permutation feature importance]{\label{fig:permutation-importance}    \includegraphics[scale=0.225]{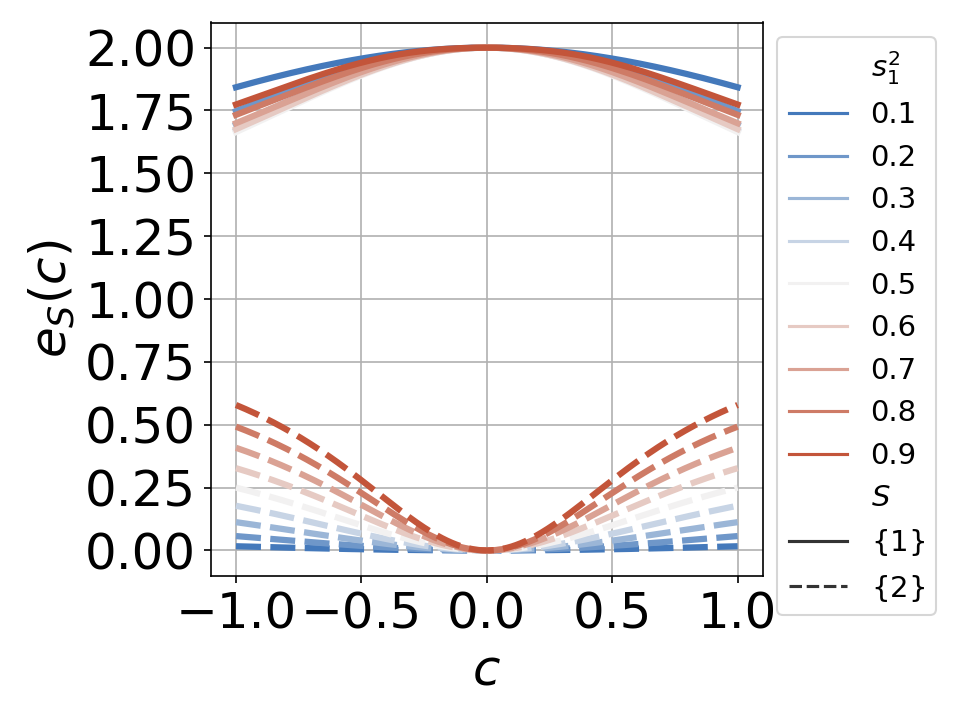}}
    \caption{Analytical approximations of faithfulness and permutation feature importance. Shown is a family of curves as a function of feature correlation $c \in [-1, 1]$ variance $s_1^2$ for constant variance $s_2^2=0.5$. Importance maps differ in offsets, indicating consistently higher importance for the informative feature $x_1$.  Yet, both methods allocate importance also to the suppressor feature $x_2$ for $c > 0$. Analogous figures for different $s_2^2$ values are contained in the supplementary Figures \ref{fig:appendix-faithfulness} and \ref{fig:appendix-permutation}.
}
    \label{fig:faithfulness-permutation-importance}
\end{figure}
It is widely acknowledged that the correctness of any XAI method as well as the correctness of a given importance map is notoriously hard to assess. This is, because there exists no agreed upon definition of importance as well as because `true' importances scores are rarely available when it comes to solving problems with learning algorithms. Nonetheless, surrogate metrics have been defined to work around this problem. These metrics are often referred to as `faithfulness' and, rather than being based on fundamental properties of the data and/or model, they are often based on predictability arguments. Faithfulness is not a well-defined concept and has numerous notions, some of which are tied to specific XAI methods~\cite{jacoviFaithfullyInterpretableNLP2020}. As these metrics are often defined algorithmically, they can be regarded as XAI methods in their own right. 

The most widely adopted notion of faithfulness is that the omission or obfuscation of an important feature will lead to a decrease in a model's prediction performance. One algorithmic operationalization to assess this is the `pixel flipping' method~\cite{samekEvaluatingVisualizationWhat2015}. 
For linear models, the simplest form of flipping or removing features is just by setting their corresponding weights $w_j$ to zero. With this, we can  approximate the classification losses through squared errors as
\begin{equation}
    \label{eq:faithfulness}
    \begin{split}
        e_{S}(\x)& \coloneqq \Eb{(Y-f_{w_{S}=0}(\x))^2} - \Eb{(Y-f(\x))^2} \;.
    \end{split}
\end{equation}
For features $x_1$ and $x_2$, we obtain
\begin{equation}
    \label{eq:faithfulness-x1-x2}
    \begin{split}
          e_{\{1\}}(\x)
           &= 2 \alpha - \alpha^2 + \alpha^2 s_1^2 ( 2c^2 -1), \\
          e_{\{2\}}(\x)
        &= \alpha^2  c^2 s_1^2 \;,
    \end{split}
\end{equation}
as derived in Appendix \ref{app:faithfulness}.
We can observe that for non-zero correlation $c$, $e_{\{2\}}$ is non-zero; that is, pixel-flipping assigns importance to the suppressor feature $x_2$. 
%

%
\subsection{Permutation Feature Importance}
\label{subsec:permutation-feature-importance}
Proposed by \citet{breimanRandomForests2001}, the permutation feature importance (PFI) for features $x_S$ measures the drop in classification performance when the associations between $x_S$ and the corresponding class labels is broken via random permutation of the values of $x_S$. As in pixel flipping, a significant drop in performance defines an important feature (set). Let $\pi_S(\x)$ be the randomly permuted version of $\x$, where features with indices in $S$ are permuted and the remaining components are untouched.
The randomly permuted features $\pi_S(\x)$ and $x_S$ are independent and identically distributed now, which leads to the following approximation of PFI:
\begin{equation}
    \label{eq:permutation-feature-importance}
    \begin{split}
        e_{S}(\x) & \coloneqq \Eb{(Y-f(\pi_S(\x)))^2} - \Eb{(Y-f(\x))^2} \;.
    \end{split}
\end{equation}
For features $x_1$ and $x_2$, we obtain
\begin{equation}
    \label{eq:permutation-feature-importance-x1-x2}
    \begin{split}
    e_{\{1\}}(\x)
        = 2 \alpha + 2 \alpha^2 c^2 s_1^2 \quad
    e_{\{2\}}(\x)
        = 2 \alpha^2 c^2 s_1^2 \;.
    \end{split}
\end{equation}
Thus, similar to faithfulness, PFI assigns non-zero importance to $x_2$ if $|c| > 0$. This similarity is expanded upon in Appendix \ref{app:pfi}, and a graphical depiction of that behavior is presented for both methods in Figure~\ref{fig:faithfulness-permutation-importance}.

\subsection{Partial Dependency Plots}
\label{subsec:partial-dependency-plots}
\begin{figure}
\centering
    \includegraphics[scale=0.2]{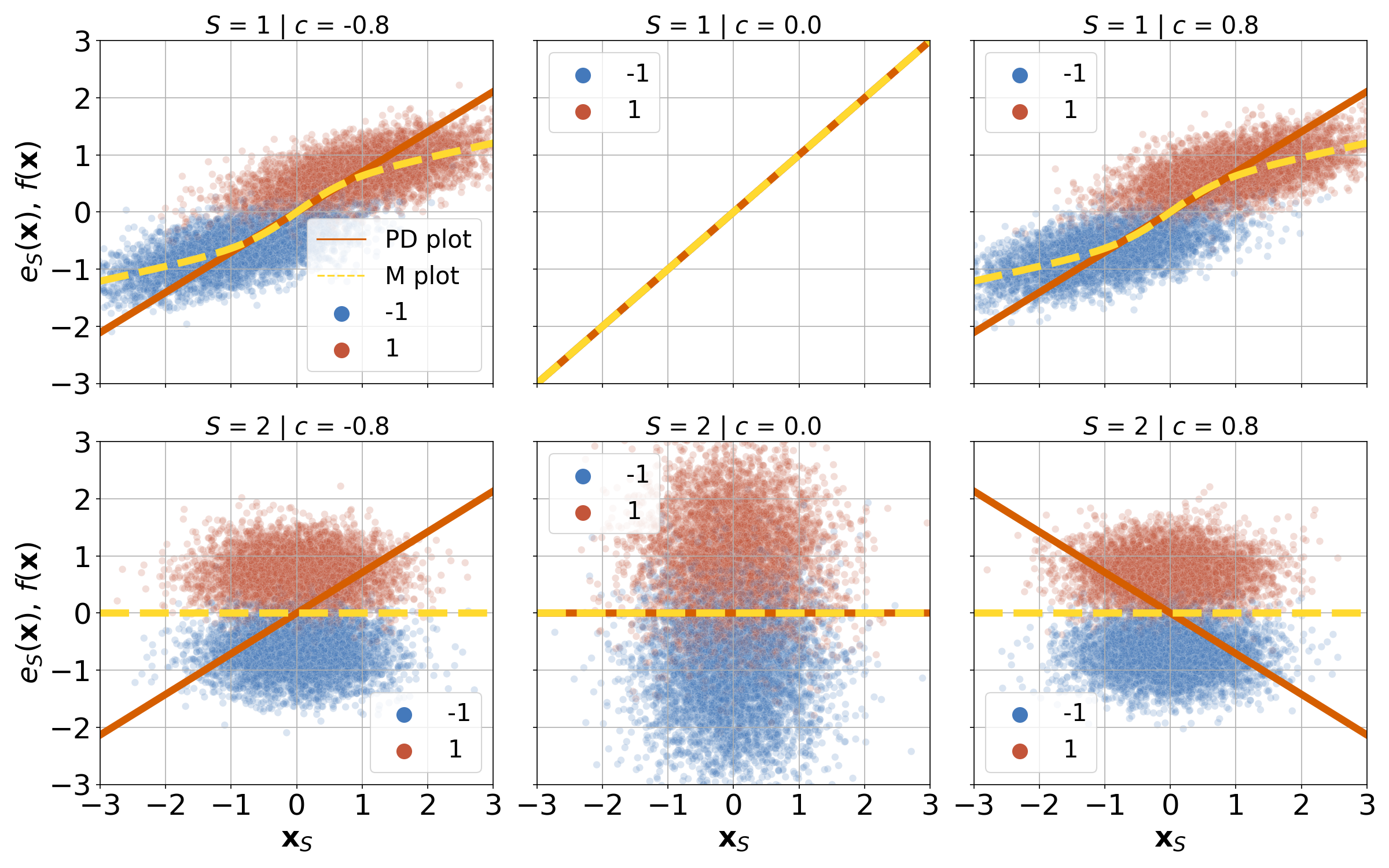}
\caption{The Partial Dependency Plots (black solid line) and M-plots (red dashed line) for different correlations (columns), and different features $x_1$ (upper row) and $x_2$ (bottom row) corresponding for Figure {fig:1}. The background shows a scatter plot of the corresponding predictions $f(\x)$ vs. the feature of interest $\x_{S}$. The Partial Dependency Plots and M-plots both `follow' the `trend’ of the samples showing an apparent dependency on the feature $x_1$ (upper row). For feature $x_2$ the scatter plots show no structural direction, where we would suspect no `directional response’ from explanation methods like those shown by the M-plots. While PD plots show a dependency on $x_2$.
The figures depict cropped versions; the scatter plots and explanation functions extend beyond the axes' limits for some plots.
}
\label{fig:partial-dependency-plots}
\end{figure}
Partial dependency (PD) plots are a visualization tool for (learned) high-dimensional functions, aiming to foster a deeper understanding of the relations between their in- and outputs. 
PD plots also became widely appreciated in the XAI community, where they have been proposed as model-agnostic `interpretation' or `explanation' tools \citep[e.g.,][]{molnarInterpretableMachineLearning}. 
For a group of features of interest $x_S$ and remaining features $x_C$, the partial dependency function is the average function
\begin{equation}
    \label{eq:partial-dependency-function}
    e_{S}(\x) \coloneqq \E_{x_{C}}\bb{f(\x)} = \int_{\R}f(x_{S}, x_{C}) p(x_{C}) \mathrm{d} x_{C} \;,
\end{equation}
where $p(x_{C})$ denotes the marginal probability density function, or `marginal expectation', of $x_{C}$.
The Bayes optimal decision \eqref{eq:bayes-optimal-weights} allows us to directly state the partial dependency functions for features $x_1$ and $x_2$ as
\begin{equation}
    \label{eq:partial-dendency-suppressor}
    \begin{split}
        e_{\{1\}}(\x) =  \alpha x_1 \quad
        e_{\{2\}}(\x) = -\alpha c s_1 s_{2}^{-1}x_2 \;.
    \end{split}
\end{equation}
%
These results indicate that the PD function does vary as a function of the suppressor feature $x_2$. This is further illustrated in Figure~\ref{fig:partial-dependency-plots}, which shows PD plots with corresponding scatter plots of the log odds $f(\x)$ as a function of  the feature of interest $x_{S}$. The partial dependency function for $x_2$ is heavily influenced by the correlation of $x_1$ and $x_2$ and only vanishes for $c=0$, indicating that PD plots are indeed merely a tool to visualize relations between in- and outputs of a function rather than providing `explanations' compatible with the data-driven definition of feature importance \eqref{eq:importance_definition}. This is in line with works reporting problematic behavior of PD plots when applied to strongly correlated data \citep{apleyVisualizingEffectsPredictor2019, molnarInterpretableMachineLearning}.
\paragraph{Marginal Plots}
For exploratory analyses of tabular datasets, it is common to start by visually assessing simple scatter plots of the target variable as a function of individual features.
As such, it is common to fit curves to pairs of in- and outputs $(x_1, y)$ and $(x_2, y)$. 
This can be done by estimating the conditional expectations $\Eb{Y | X_1 = x_1}$ or $\Eb{Y | X_2 = x_2}$. 
A variation of this is to replace output parameters by their model predictions, leading to conditional expectations $e_{S}(\x) \coloneqq \Eb{f(x_{S}, x_{C})| X_{S} = x_{S}}$, which were coined M-plots by \citet{apleyVisualizingEffectsPredictor2019}. 
Their Calculation requires the conditional expectations $\Eb{X_2|X_1=x_1} = \frac{c s_2}{s_1} h(x_1)$ and $\Eb{X_1|X_2=x_2} = \frac{c s_1}{s_2} x_2$,
%
%
where
\begin{equation}
    \label{eq:h-function}
    h(x_1) \coloneqq (x_1 - 1) \vartheta(\nicefrac{2 x_1}{s_1^2}) 
    + (x_1 + 1) (1 - \vartheta(\nicefrac{2 x_1}{s_1^2}))
\end{equation}
%
and with $\vartheta(x) \coloneqq (1 + \exp(-x))^{-1}$ as the sigmoid function.
For the generative model \eqref{eq:suppressor-problem} and corresponding Bayes optimal classifier with weights \eqref{eq:bayes-optimal-weights} , the conditional expectations for the model given $x_1$ or $x_2$, respectively, amount to
\begin{equation}
    \label{eq:m-plot-functions}
    \begin{split}
        e_{\{1\}}(\x)
        &= \alpha x_1 - \alpha c^2 h(x_1) \quad
        e_{\{2\}}(\x) = 0 \;.
    \end{split}
\end{equation}
This is shown in Appendix \ref{app:cond-expectations}.
Thus, the M-plot assigns a vanishing conditional expectation value to the suppressor variable $x_2$, which is also confirmed visually in Figure~\ref{fig:partial-dependency-plots} (bottom row). As such, M-plots appear to be suitable tools to identify important features according to definition \eqref{eq:importance_definition}. However, M-plots have been reported to lead to misinterpretations of main effects if $y$ depends on $x_1$ and $x_2$, especially when there is an interaction between the two features~\cite{grompingModelagnosticEffectsPlots2020}. Studying the case of interacting features, however, goes beyond the scope of this paper.

%

%
\subsection{Shapley Values}\label{subsec:shapley-values}
Another class of XAI methods leverages game theoretic considerations to assign importance scores to individual features. 
Originally introduced by \citet{shapleyValueNPersonGames1953}, the concept of distributing gains of a coalition game among players fairly was extended by
\citet{lipovetskyAnalysisRegressionGame2001} and \citet{ lundbergUnifiedApproachInterpreting2017}, who propose the use of Shapley values~\cite{shapleyValueNPersonGames1953} as a procedure to quantify the contribution of a feature to a decision function by considering all possible combinations of features. 
One can quantify the contribution of a feature $x_j$ to a coalition of features $S$ via the Shapley value
\begin{equation}
    \label{eq:shapley-value}
    e_{\{j\}} = \sum_{S \subseteq [d] \setminus \{j\}}
    \gamma_{d}(S) \left[ v(S \cup \{j\}) - v(S) \right] \;,
\end{equation}
with the weighting factor $\gamma_d$ representing the proportion of coalitions $S$ not including the $j$th feature, defined as $\gamma_{d}(S) = \nicefrac{|S|! (d - |S| - 1)!}{d!}$. 
The value function $v: 2^{[d]} \to \R$, with $v(\emptyset) = 0$, is a set function that assigns a quantity of `worth' to a coalition and can have many forms.
But, for our analysis, we are focusing on the choices made by \citet{lipovetskyAnalysisRegressionGame2001, lundbergUnifiedApproachInterpreting2017} and \citet{aasExplainingIndividualPredictions2020}.
In general, the purpose of the value function  $v(S)\coloneqq g_S(\x_S)$, $g_S: \R^{|S|} \to \R$ is to measure the impact of a reduced subset of feature values $x_S$ on the model output.
In the following paragraphs, we analyze three different value functions to assess: (1) their impact on feature attribution within the Shapley value framework, and (2) the consequences for models relying on suppressor variables.

\paragraph{Coefficient of Multiple Determination}
\label{par:coef-determination}
In the Shapley value regression context, \citet{lipovetskyAnalysisRegressionGame2001} leverage the coefficient of determination~\cite{hoffmanParamorphicRepresentationClinical1960} as a value function, which we decompose as $R^2 = \sum_{j = 1}^{d} w_j r_j$.
Here, $w_j$ are the learned model weights, and $r_j \coloneqq (X^{\top}y)_j$ defines the sample correlation between feature $x_j$ and target $y$, for standardized features $x_j$.
We can directly define $R^2$ for a subset of features as $g_S(\x_S) \coloneqq  R_S^2 = \sum_{j \in S} w_j r_j$,
%
%
and utilize it as value function $v(S)\coloneqq g_S(x_S)$, which can be interpreted as shares of the overall $R^2$.
If we recall the data generation process \eqref{eq:suppressor-problem} and consider the covariances $\mathrm{Cov}(Y, X_1) = 1$, and $\mathrm{Cov}(Y, X_2) = 0$, respectively, we can state the marginal Pearson correlations $\rho_{Y, X_1} = (s_1^2 + 1)^{-1/2}$ and $\rho_{Y, X_2} = 0$ directly, without relying on the sample correlations $r_j$. 

First, we consider the  case of calculating the Shapley values $e_{\{j\}}$ with respect to the $R_S^2$ value function, and, as originally intended by \citet{lipovetskyAnalysisRegressionGame2001}, three hypothetically trained models: One bivariate model, here the Bayes rule \eqref{eq:bayes-optimal-weights}, and two univariate models  $f_{\{1\}}(\x) = \hat{w}x_1$ and $f_{\{2\}}(\x)=\tilde{w}x_2$. We specify $e_{\{1\}}$, $e_{\{2\}}$ as
%
%
%
\begin{equation}
    \label{eq:shapey-values-2d-general}
    \begin{split}
        e_{\{1\}}(\x)
        = \frac{\alpha + 1}{2 (s_1^2 + 1)^{1/2}} \quad 
        e_{\{2\}}(\x)
        = \frac{\alpha - 1}{2(s_1^2 + 1)^{1/2}},
    \end{split}
\end{equation}
where the rules $f_{\{1\}}(\x)= x_1$ and $f_{\{2\}}(\x) = x_2$, with $\hat{w}=1$ and $\tilde{w}=0$ correspond to the optimal decisions for the univariate models. 
%
We can observe that the Shapley values are `governed' by the factor $\alpha$ of the bivariate model. 
As long as $c \neq 0$, it holds that $\alpha \neq 1$, and this method attributes importance to the suppressor feature $x_2$.
Now, we approximate this procedure using only the bivariate model containing all variables -- this is the `common' scenario, as it can be quite computationally expensive to train new models on many feature subsets.
Using the Shapley value framework together with the $R^2$ measure, we obtain
\begin{equation}
    \label{eq:shapey-values-bayes-one-model}
    \begin{split}
        e_{\{1\}}(\x) &= \alpha (s_1^2 + 1)^{-1/2}, \quad
        e_{\{2\}}(\x) = 0 \;.
    \end{split}
\end{equation}
Since $e_{\{2\}} = 0$, we can conclude that $R^2$ measure in combination with Shapley values is an appropriate value function for assessing feature importance for our linear data generation process \eqref{eq:suppressor-problem}. This, and the work of the following section, is expanded upon in Appendix \ref{app:shapley-values}.

\paragraph{SHAP}
\label{par:shap}
\citet{lundbergUnifiedApproachInterpreting2017} propose the conditional expectation for a suitable approximation of $f$, but for computational reasons the authors decided to approximate it with the non-conditional expectation, assuming feature independence. This is called the SHAP (Shapley additive explanations) approach.
Later, \citet{aasExplainingIndividualPredictions2020} suggested an estimation method for the conditional expectation, extending SHAP by actively incorporating potential dependencies among features.
We start by defining the value function via the marginal expectation $g_S(\x_S)\coloneqq\E_{x_{C}}\bb{f(x_{S}, x_{C})}$, and with the results of Section~\ref{subsec:partial-dependency-plots}, we obtain the Shapley values
\begin{equation}
    \label{eq:shapey-values-marginal-expectation}
    \begin{split}
        e_{\{1\}}(\x) = \alpha x_1,  \quad
        e_{\{2\}}(\x) = -\alpha c s_1 s_2^{-1} x_2 \;.
    \end{split}
\end{equation}
This, in essence, resembles the partial dependency functions \eqref{eq:partial-dendency-suppressor}. 
In a similar way, we calculate the Shapley values for the set function defined via the conditional expectation $g_S(\x_S)\coloneqq\Eb{f(x_{S}, x_{C})| X_{S} = x_{S}}$ as
\begin{equation}
    \label{eq:shapey-values-conditional-expectation}
    \begin{split}
        e_{\{1\}}(\x) &= \alpha x_1 - \frac{\alpha c^2}{2} h(x_1) - \frac{\alpha c s_1}{2 s_2} x_2 \\
        e_{\{2\}}(\x) &= \frac{\alpha c^2}{2} h(x_1) - \frac{\alpha c s_1}{2 s_2} x_2 \, ,
    \end{split}
\end{equation}
%
where $h$ is defined in \eqref{eq:h-function}.
Thus, the Shapley value $e_{\{2\}}$ does not just reflect an attribution of importance to the suppressor variable $x_2$ but is also affected by feature $x_1$ if $c \neq 0$.

\subsection{Counterfactual Explanations}

\begin{figure}
        \centering
        \includegraphics[width=0.25\textwidth]{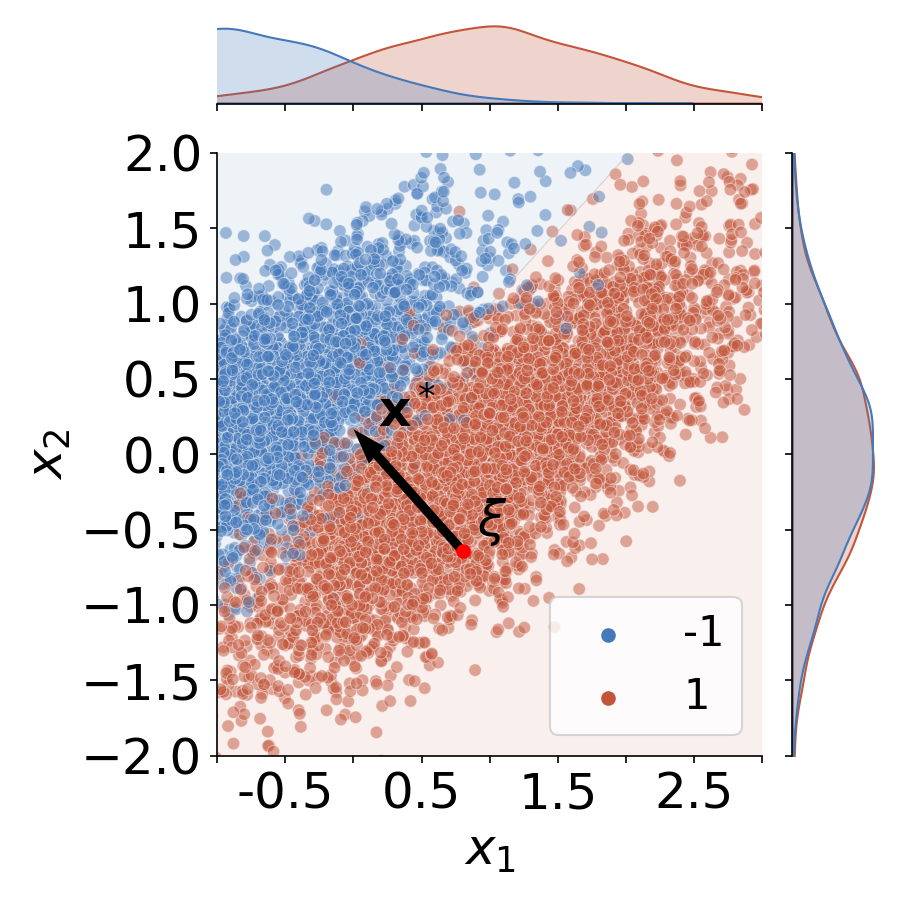}
        \caption{Counterfactual $x^*$ for a given instance of interest $\xi$ in the generative setting $c=0.8$, $s_1^2=0.8$, and $s_2^1=0.5$. As can be seen, for $|c| > 0$, reaching a counterfactual decision always involves a manipulation of the suppressor feature $x_2$.
                \label{fig:counterfactuals}
}
\end{figure}

\citet{wachterCounterfactualExplanationsOpening2017} propose an explanation framework based on counterfactual explanations, which we can think of as statements depicting an ``alternative world''. 
Formally, we have a given instance $\xi \in \R^d$ and the desired outcome $y^*$, and try to find a minimizer
\begin{equation}
    \label{eq:counterfactual-problem}
    \x^* = \text{arg}\min_{\x} \, \max_{\lambda} \, \lambda (f(\x) - y^*)^2 + \delta(\x, \xi) \;,
\end{equation}
for $\lambda \in \R$ and a suitable distance function $\delta$~\cite{wachterCounterfactualExplanationsOpening2017}.
To find a counterfactual sample according to \eqref{eq:counterfactual-problem} for our linear model $f(\x) = w^{\top} \x$, it is sufficient to consider points that are located on the linear decision boundary $f(\x^*)=0$ of the Bayes optimal classifier \eqref{eq:bayes-optimal-weights}, since the decision can be flipped in any epsilon-neighborhood around any such point. The closest such counterfactual $\x^*$ for a given instance $\xi$ is the point that has minimal distance to $\xi$ in the Euclidean sense. We can also think of that point as the orthogonal projection of $\xi$ onto the decision hyperplane via its orthogonal subspace
\begin{equation}
    \label{eq:counterfactual-orthogonal-projection}
    \langle \xi - a u, u \rangle = 0 \quad \text{with} \quad \x^* \coloneqq \xi - a u \;,
\end{equation}
where $u$ is an element of the orthogonal complement of $w$, and $a \in \R$.
Then, with $u = (c s_1 /s_2, 1)^{\top}$ and $a = \langle \xi, u\rangle / \|u\|_{2}^{2}$, the counterfactual explanation $\x^*$ results in
\begin{equation}
    \label{eq:counterfactual-linear-model}
    \begin{split}
            x_1^* &= \beta (\xi_1 - \xi_2 c s_1 s_2^{-1}) \\
            x_2^* &= \beta c s_1 s_2^{-1} (\xi_2 c s_1 s_2^{-1} + \xi_1) \;,
    \end{split}
\end{equation}
with $\beta \coloneqq ((c s_1 s_2^{-1})^2 + 1)^{-1}$.
Thus, to change the decision of the Bayes optimal classifier with minimal interventions, a shift from $\xi$ to $\x^*$ would be required, and this shift would not only involve a change in the informative feature $x_1$ but also in the suppressor feature $x_2$ (see also Figure~\ref{fig:counterfactuals} for a graphical depiction). Based on this result it may be, erroneously, concluded from this counterfactual explanation, that feature $x_2$ has a correlation with or even a causal influence on the classifier decision.
%
\subsection{FIRM}\label{subsec:firm}
Another post-hoc method to assess the importance of features of an arbitrary function $f:\R^d \to \R$ is the feature importance ranking measure (FIRM) proposed by \citet{zienFeatureImportanceRanking2009}.
Inspired by the feature sensitivity measure of \citet{friedmanGreedyFunctionApproximation2001}, the authors utilize the conditional expectation $\Eb{f(\x)|X_S = x_S}$ and define the importance ranking measure as
\begin{equation}
    \label{eq:firm}
    \begin{split}
        e_{S}(\x)
        &\coloneqq \var(\Eb{f(\x)|X_S})^{\frac{1}{2}} \;.
    \end{split}
\end{equation}
Computing this expression, in general, is infeasible since we need access to the data distribution.
For the generative model \eqref{eq:suppressor-problem} it is possible to prove that
\begin{equation}
    \label{eq:firm-x1-x2}
    \begin{split}
        e_{\{1\}}(\x) &= \alpha \var( X_1 - c^2 h(X_1))^{\frac{1}{2}} \\
        &\ge \frac{\alpha}{2} \left(2 \vartheta(2/s_1^2) - 1 \right) \\
        e_{\{2\}}(\x) &= 0 \;.
    \end{split}
\end{equation}
A derivation of the lower bound is provided in Appendix \ref{app:firm}.
As also noted in \citet{haufeInterpretationWeightVectors2014}, the variability of $e_{\{2\}}$ is zero, indicating that FIRM does not assign importance to suppressor features.

\subsection{Integrated Gradients}\label{subsec:integrated-gradients}
Integrated gradients~\cite{sundararajanAxiomaticAttributionDeep2017} belongs to the family of path methods~\cite{friedmanPathsConsistencyAdditive2004}, which aggregate a model's gradients along a predefined path or curve $\gamma: [0, 1] \to \R^d$ with $\gamma(0) = \x'$ and $\gamma(1)=\x$.
If we think of images, then $\x \in \R^d$ can be an image we seek an explanation for, and $\x'$ represents a corresponding baseline image, where a black image $\x' \equiv 0$ is a common choice.
For the curve $\gamma: t \mapsto \x' + t(\x-\x')$, a general baseline $\x'$, and a model $f$, the integrated gradient importance map is given by \citep{sundararajanAxiomaticAttributionDeep2017}
\begin{equation}
    \label{eq:integrated-gradients}
    \begin{split}
        e_{\{j\}}(\x) &\coloneqq (x_j - x'_j) \int_{[0,1]} \frac{\partial f(\x' + t(\x-\x'))}{\partial x_j}  \,\mathrm{d}t.
    \end{split}
\end{equation}
For the Bayes optimal linear classifier \eqref{eq:bayes-optimal-weights}, the importance scores for features $x_1$ and $x_2$ are given by
\begin{equation}
    \label{eq:integrated-gradients-x1-x2}
    \begin{split}
            e_{\{1\}}(\x) &= \frac{\alpha}{2} (x_1^2 - (\x')^2), \\
            e_{\{2\}}(\x) &= - \frac{\alpha c s_1}{2 s_2} (x_2^2 - (\x')^2) \;,
    \end{split}
\end{equation}
respectively. Thus, independent of the baseline $\x'$ (provided that $\x' \neq \x$), the integrated gradients for the suppressor feature $x_2$ are non-zero for $|c| > 0$.

\subsection{LIME}\label{subsec:lime}
The idea of LIME~\cite{ribeiroWhyShouldTrust2016} is to `explain' a model's decision for a given instance $\x$ by sampling data points in the vicinity of $\x$ and using these samples to build a `glass-box' model, which is assumed to be more easily interpretable. Typically, a linear model is chosen as a surrogate model. In the scenario studied here, the Bayes optimal model \eqref{eq:bayes-optimal-weights} is already linear with non-zero weight $w_2$. 
Thus, we would expect that a local linear approximation would show the same behavior. Indeed, \citet{garreauExplainingExplainerFirst2020} show that for a `linear black-box' model and a Gaussian \textit{i.i.d.} sampling procedure from $N(\mu, \sigma^2 \mathrm{I}_d)$, the local weights $\hat{w}_j$ estimated by LIME are approximately proportional to the partial derivatives of $f$. Since these derivates reduce to the weights \eqref{eq:bayes-optimal-weights} of the Bayes optimal linear classifier in the studied setting, we have $w_j \propto \hat{w}_j$. Therefore, LIME resembles the global model and attributes non-zero importance to the suppressor variable $x_2$.
\subsection{Saliency Maps, LRP and DTD}\label{subsec:saliency-methods}

Saliency map explanations estimate how a prediction $f(\x)$ is influenced when moving along a specific direction in the input space. If the direction is along the model's gradient, this is known as sensitivity analysis \cite{baehrensHowExplainIndividual2010, Simonyan14a}.
Several explanation techniques for neural networks are based on this approach (e.g. DeConvNet and Guided BackProp), primarily distinguishing themselves by their treatment of rectifiers \cite{kindermansLearningHowExplain2017, zeilerVisualizingUnderstandingConvolutional2014, springenbergStrivingSimplicityAll2015}.
For single-layer neural networks without rectifiers, that is, linear models, the saliency maps of these explanation methods reduce to the gradient itself (cf. Section \ref{subsec:gradient}).
Layerwise relevance propagation \citep[LRP,][]{bachPixelWiseExplanationsNonLinear2015} and its generalization Deep Taylor Decomposition \citep[DTD,][]{montavonExplainingNonLinearClassification2017} are methods that propagate a quantity termed `relevance' from output to input neurons backwards through a neural network, following a set of rules. The DTD approach develops, for each layer $l$ of a neural network, a first-order Taylor expansion around a root point $\x_0$, which gives rise to a relevance score for each neuron $j$ with the propagation rule
$e_{\{j\}}(\x) \coloneqq R_{j}^{l-1} = w \odot (\x-\x_0) (w^{\top}\x)^{-1} R_{j}^{l}$
, where $\odot$ is the Hadamard product.
Choosing an appropriate root point is essential in the DTD framework, and \citet{kindermansLearningHowExplain2017} notice that by estimating the distractor $\eta$ and understanding it as root point $x_0=\eta$, DTD recovers the pattern estimator for linear models proposed by \citet{haufeInterpretationWeightVectors2014}. 
 \citet{kindermansLearningHowExplain2017} derive the signal estimator $S_{\abf} =  \operatorname{Cov}(\x, y) w^{\top} \x$, yielding the DTD propagation rule
 %
 \begin{equation}
     \label{eq:pattern-attribution}
     e_{\{j\}}(\x) = (w \odot \abf)_j
 \end{equation}
for $j=1,2$ (cf. Eg. \eqref{eq:suppressor-problem}) (see Appendix~\ref{app:lrp}). \citet{kindermansLearningHowExplain2017} refer to this propagation rule as \textit{PatternArribution}, or \textit{PatterNet} in case where only the activation patterns $\abf_j$ are back-propagated.
In this case, DTD indeed achieves that no relevance gets attributed to suppressor features in a linear setting.
Notably, it has also been shown that in more complex learning scenarios and depending on root points, 
DTD can generally yield almost any explanation \cite{kohlbrenner2020TowardsBestPraticeExplNN, montavon2018MethodsInterpretingNN, sixt2022RigorousStudyDeepTylor}. 

\section{Discussion}

The field of XAI is seen as a critical part of a to-be-developed infrastructure that should guarantee the safety of future ML-based high-stake decision systems and create trust in such systems.
However, the current state of XAI lacks precise specifications of the problem to be solved by XAI methods. Operationalizations of XAI are, therefore, notoriously difficult to validate theoretically and empirically, which currently prohibit their use for quality assurance.

Two proclaimed use cases of XAI are model and dataset debugging \cite{lapuschkinUnmaskingCleverHans2019}, and feature discovery \citep[e.g.][]{jimenezLunaDrugDiscoveryExplainable2020, tranDeepLearningCancer2021}. However, it remains unclear how well contemporary XAI methods can provide evidence in each of these use cases that is beyond anecdotal. If XAI outputs are ill-defined or simply unfit for purpose, this could turn an anticipated benefit of their use even into a disadvantage. For example, characterizing features as important that have no statistical association with the prediction target could give rise to psychological biases and circular reasoning. A pathologist presented with an importance or saliency map for a histological image may try to identify familiar patterns in the map while potentially being tempted to ignore less familiar structures. In the worst case, this could mutually reinforce false prior beliefs between researchers and developers of XAI methods.

\subsection*{Suppressors as Benchmarks for XAI}
\citet{wilmingScrutinizingXAIUsing2022} argue that humans often implicitly assume an actual statistical association between a feature and the prediction target when being offered the `explanation’ that the feature in question is important. This gives rise to a purely data-driven yet concrete definition of feature importance based on a statistical dependency on the prediction target.
We use this definition to construct a standard binary classification problem with Gaussian class-conditional distributions. By introducing noise correlations within this model, we create a suppressor variable, which has no statistical relation to the target but whose inclusion in any model will lead to better predictions \citep{haufeInterpretationWeightVectors2014}.

We view this simple, yet very insightful, classification problem primarily as a minimal counterexample, where the existence of suppressor variables challenges the assumptions of many XAI methods as well as the assumptions underlying metrics such as faithfulness, which are often considered a gold-standard for quantitative evaluation and an appropriate surrogate for `correctness'. Indeed, authors have shown empirically that XAI methods can lead to suboptimal `explanation performance' even when applied to linear data with suppressor variables \citep{wilmingScrutinizingXAIUsing2022}. 
Here, we complement the study of \citet{wilmingScrutinizingXAIUsing2022} by deriving analytical expression of popular XAI methods employing a two-dimensional linear binary classification problem that has the same problem structure as the 64-dimensional problem presented by \citet{wilmingScrutinizingXAIUsing2022}.
These analytical expressions allow us to study the factors that lead to non-zero importance attribution, and to expose the mathematical mechanism by which different properties of the data distribution influence XAI methods.
Our results demonstrate that outputs of explanation methods must be interpreted in combination with knowledge about the underlying data distribution.
Conversely, it may be possible that XAI methods with improved behavior could be designed by reverse-engineering the analytical importance functions $e_{S}$.

We found that several XAI methods are incapable of nullifying the suppressor feature, i.e., assigning non-zero importance to it, when correlations between features are present. 
This is the case for the naive pixel flipping and the PFI methods representing operationalization of faithfulness, but also for actively researched methods like SHAP, LIME, and counterfactuals, as well as partial dependency plots. 
Note that these methods can typically also not be `fixed' by just ranking features according to their importance scores and considering only the top features `important'. In fact, we can devise scenarios where the weight $w_2$ corresponding to the suppressor variable $x_2$ is more than twice as high as the weight $w_1$ (see Appendix \ref{app:ranking} and \citet{haufeInterpretationWeightVectors2014}), which may lead to the misconception that the feature $x_2$ is `twice' as important as feature $x_1$.
XAI methods based on the Shapley value framework yield particular diverging results, 
as the strong influence of the value function is reflected in the diversity of analytical solutions.
SHAP-like approaches, based on the conditional or marginal expectations \ref{subsec:shapley-values}, show how heavily dependent such methods are on the correlation structure of the dataset.
In contrast, the M-Plot approach, FIRM, PATTERN, and the Shapley value approach using the $R^2$ value function, deliver promising results by assigning exactly zero importance to the suppressor variable. This positive result can be attributed to the fact that all methods make explicit use of the statistics of the training data including the correlation structure of the data. This stands in contrast to methods using only the model itself to assign importance to a test sample.

\subsection{Limitations}
Here we studied a linear generative model and used a univariate data-driven definition of feature importance to design our ground truth data. 
In real-world scenarios, we do not expect that suppressor variables are always perfectly uncorrelated with the target.
In Appendix \ref{app:bayes-classifier} we provide deliberations for the case where the suppressor variable $x_2=\varepsilon z + \eta_2$ consists of a small portion $\varepsilon \in \R$ of the signal $z$ as well.
However, in this case, it is not exactly clear what numerical value for the importance we can assume as ground-truth, other than zero.
Furthermore, modern machine learning model architectures excel in dealing with highly complex non-linear data involving, among other characteristics, feature interactions.  Most XAI methods have been designed to `explain' the predictions of such complex models. To better understand the behavior of both machine learning models and XAI methods in such complex settings, future work needs to focus on non-linear cases, and develop clear definitions of feature importance in complex settings. 

\section{Conclusion}
We study a two-dimensional linear binary classification problem, where only one feature carries class-specific information. The other feature is a suppressor variable carrying no such information yet improving the performance of the Bayes optimal classifier. Analytically, we derive closed-form solutions for the outputs of popular XAI methods, demonstrating that a considerable number of these methods attribute non-zero importance to the suppressor feature that is independent of the class label. We also find that a number of methods do assign zero significance to that feature by accounting for correlations between the two features. This signifies that even the most simple multivariate models cannot be understood without knowing essential properties of the distribution of the data they were trained on.

\section*{Acknowledgements}
This result is part of a project that has received funding from the European Research Council (ERC) under the European Union’s Horizon 2020 research and innovation programme (Grant agreement No. 758985), the German Federal Ministry for Economy and Climate Action (BMWK) in the frame of the QI-Digital Initiative, and the Heidenhain Foundation. We thank Jakob Runge for a fruitful discussion.

\bibliography{xai_better, more-references}
\bibliographystyle{icml2023}

\newpage
\appendix
\onecolumn

\section{Bayes optimal classifier}\label{app:bayes-classifier}
The generative data model, defined in \eqref{eq:suppressor-problem}, induces a binary classification problem yielding two class-conditional densities which are both multivariate normal, with $\mathbf{X} \mid Y=y \sim N(\mu_i, \Sigma)$ for $y \in \{-1, 1\}$ and $i=1, 2$, and have identical covariance matrix $\Sigma \in \R^{2 \times 2}$ and expectations $\mu_1 = (1, 0)^{\top}$ and $\mu_2 = (-1, 0)^{\top}$.
We solve the classification task in a Bayes optimal way if we assign $\x$ either to class $Y=1$ or to class $Y=-1$ based on the minimal squared Mahalanobis distance $\delta^2(\x, \mu_i) = (\x-\mu_i)^{\top} \Sigma^{-1} (\x-\mu_i)$ between $\x$ and the two class means $\mu_i, i=1,2$. 
As described we have equal covariance matrices $\Sigma$ for both classes, thus, the Bayes rule becomes linear and we can assign $\x$ to class $Y=1$, if $\w^{\top} (\x - \mu) \geq 0$, where $\w \coloneqq \Sigma^{-1} (\mu_1 - \mu_2)$ and $\mu \coloneqq \frac{1}{2} (\mu_1 + \mu_2)$. 
The concrete form of the Bayes optimal rule $f(\x) = \w^{\top}\x + b$ with weights $\w^{\top}=(w_1, w_2)^{\top}$ is determined by the coefficients
\begin{equation}
    \label{eq:app-bayes-optimal-weights}
    \begin{split}
        w_1 &= \alpha, \quad w_2 = -\alpha c s_1/s_2\\
    \end{split}
\end{equation}
for $\alpha \coloneqq (1+(cs_1/s_2)^2)^{-\frac{1}{2}}$ and $||w||_2=1$ and $b=0$.
The inverse of the covariance matrix $\Sigma$ is given by
\begin{equation}
    \label{eq:app-noise-covariance}
    \Sigma^{-1} = \frac{1}{s_1^2 s_2^2 (1-c^2)} \begin{bmatrix}
                 s_2^2 & -c s_1 s_2 \\ -c s_1 s_2 & s_1^2
    \end{bmatrix}.
\end{equation}
Furthermore, we consider a version of generative data model \eqref{eq:suppressor-problem} where we parameterize, via a scalar $\varepsilon \in \R$, the dependency between the suppressor variable and the target 
\begin{equation}
    \label{eq:suppressor-problem-eps}
    \begin{split}
        \x = \abf_{\varepsilon} z + \eta, \quad y = z,
    \end{split}
\end{equation}
with $Z \sim Rademacher(1/2)$, $\abf_{\varepsilon}=(1, \varepsilon)^{\top}$ and $H \sim N(\mathbf{0}, \Sigma)$.
The induces binary classification problem slightly changes with class-conditional distributions
$\mathbf{X} \mid Y = y \sim N(\mu_i, \Sigma)$ for $y \in \{-1, 1\}$ and $i=1,2$ and updated expectations $\mu_1 = (1, \varepsilon)^{\top}$ and $\mu_2 = (-1, \varepsilon)^{\top}$.
Then for the optimal Bayes rule, we yield the weights and offset
\begin{equation}
    \label{eq:app-bayes-optimal-weights-eps}
    \begin{split}
        w_1 &= \alpha, \quad w_2 = -\alpha c s_1/s_2 \quad b = \varepsilon \alpha c s_1 / s_2 \, . \\
    \end{split}
\end{equation}
%

\section{Ranking}\label{app:ranking}
Let the correlation $c=-0.8$ and the varainces be $s_1^2=1$ and $s_2^2 = 0.15$, then we yield the coefficients $w_1 \approx 0.42$ and $w_2 \approx 0.90$.
Using the coefficient as importance scores would rank the suppressor variable $x_2$ twice as high as the class-dependent variable $x_1$.
%
\section{Faithfulness}\label{app:faithfulness}
Throughout the appendix, let the random variables $\mathbf{X}=(X_1, X_2)$ and $Y$ be defined as in Section \ref{sec:methods} and $f(\x)=w_1 x_1 + w_2 x_2$.
For the Pixel-Flipping method, we consider the error
\begin{equation}
   e_{S}(\x) \coloneqq \Eb{(Y-f_{w_{S}=0}(\x))^2} - \Eb{(Y-f(\x))^2}.
\end{equation}
Now, let us consider 
\begin{equation}\label{eq:expectation-y-f}
\begin{split}
       \Eb{(Y-f(\x))^2} 
       &= \Prob(Y=1) \Eb{(Y - f(\x))^2 \mid Y =1} \\
       &+ \Prob(Y=-1) \Eb{(Y - f(\x))^2 \mid Y =-1} \\
       &= \frac{1}{2}  \Eb{(1 - f(\x))^2 \mid Y =1} + \frac{1}{2 }\Eb{(1 + f(\x))^2 \mid Y =-1} \\
       &= 1 - 2 w_1 + w_1^2 (s_1^1 + 1) + w_2^2 s_2^2 + 2 w_1 w_2 c s_1 s_2.
\end{split}
\end{equation}
We take a closer look at the conditional expectation $\Eb{(1 - f(\x))^2 \mid Y =1}$ and observe
\begin{equation}\label{eq:expectation-1-f}
\begin{split}
        \Eb{(1 - f(\x))^2 \mid Y =1} &= \Eb{1 \mid Y=1} - 2 \Eb{f(\x) \mid Y = 1} + \Eb{f(\x)^2 \mid Y = 1} \\
        &= 1 - 2 w_1 + w_1^2 (s_1^1 + 1) + w_2^2 s_2^2 + 2 w_1 w_2 c s_1 s_2,
\end{split}
\end{equation}
where 
\begin{equation}\label{eq:expectations-f-and-f2}
\begin{split}
        \Eb{f(\x) \mid Y =1} &= w_1, \\
        \Eb{f(\x)^2 \mid Y =1} &= \operatorname{Var}\big( f(\x) \mid Y=1\big) + \Eb{f(\x) \mid Y=1}^2 \\
        &= w_1^2 s_1^2 + w_2^2 s_2^2 + w_1^2 + 2 w_1 w_2 c s_1 s_2.
\end{split}
\end{equation}
By using \eqref{eq:expectation-1-f} and \eqref{eq:expectations-f-and-f2} we can compute the value for $\Eb{(1 + f(\x))^2 \mid Y =-1}$ analogously
\begin{equation}\label{eq:expectation-1+f}
\begin{split}
        \Eb{(1 + f(\x))^2 \mid Y =-1}
        &= 1 - 2 w_1 + w_1^2 (s_1^1 + 1) + w_2^2 s_2^2 + 2 w_1 w_2 c s_1 s_2.
\end{split}
\end{equation}
%
%
Similarly, we reive the results for $\Eb{(Y-f_{w_{S}=0}(\x))^2}$ using the obfuscated decision rule $f_{w_{S}=0}$.
Finally, for the weights $w_1 = \alpha$ and $w_2=\alpha c s_1 / s_2$ we yield the importance values \eqref{eq:faithfulness-x1-x2}
\begin{equation}\label{eq:faithfulness-appendix-x1-x2}
\begin{split}
        e_{\{1\}}(\x) &= \Eb{(Y-f_{w_1=0}(\x))^2} - \Eb{(Y-f(\x))^2} \\
        &= 1 + w_2^2 s_2^2 - 1 - 2 w_1 + w_1^2 (s_1^1 + 1) + w_2^2 s_2^2 + 2 w_1 w_2 c s_1 s_2 \\
        &= 2 \alpha - \alpha^2 + \alpha^2 s_1^2 ( 2c^2 -1) \\
        e_{\{2\}}(\x) &= \Eb{(Y-f_{w_2=0}(\x))^2} - \Eb{(Y-f(\x))^2} \\
        &= 1 - 2w_1 + w_1^2 (s_1^2 + 1) - 1 - 2 w_1 + w_1^2 (s_1^1 + 1) + w_2^2 s_2^2 + 2 w_1 w_2 c s_1 s_2 \\
        &= \alpha^2 s_1^2 c^2 \\
\end{split}
\end{equation}
\begin{figure}
    \centering
    \subfloat[Faithfulness for $s_2^2=0.1$]{\label{fig:faithfulness-s2-01}    \includegraphics[scale=0.4]{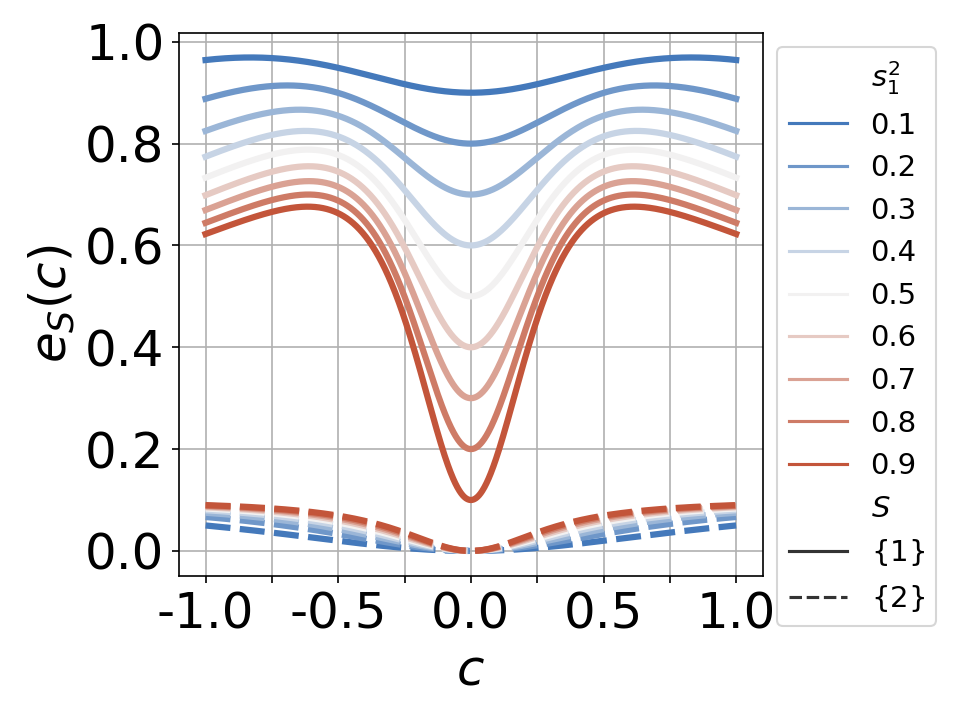}}
\hspace{0.05\columnwidth}
    \subfloat[Faithfulness for $s_2^2=0.9$]{\label{fig:faithfulness-s2-09}    \includegraphics[scale=0.4]{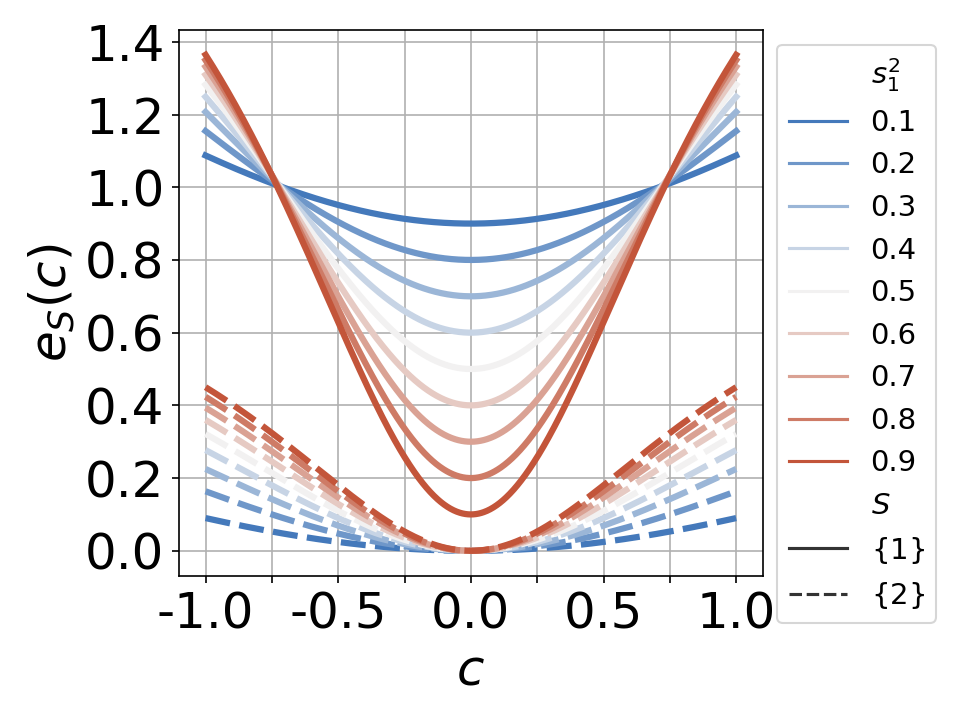}}
    \caption{}
    \label{fig:appendix-faithfulness}
\end{figure}
%

\section{Permutation Feature Importance}\label{app:pfi}
Analogously to the computation of the Faithfulness values we can compute the Permutation Feature Importance values.
Note, we compute the Permutation Importance value in a relatively naive way, where we understand the permutation $\pi_S(\x)$ as `breaking' the correlations with the remaining features and the target.
We do not provide a probabilistic definition of a permutation operator.
We just use a direct translation of how we would implement feature permutation in practice. 
We already computed the value of $\Eb{(Y-f(\x))^2}$, therefore it is sufficient to consider 
\begin{equation}\label{eq:appendix-permutation-importance-x1}
    \begin{split}
         \Eb{(Y-f(\pi_{\{1\}}(\x)))^2} 
         &= \Eb{Y^2} - 2 \Eb{Yf(\pi_{\{1\}}(\x))} + \Eb{f(\pi_{\{1\}}(\x))^2} \\
         &= 1 + w_1^2(s_1^2 + 1) + w_2^2 s_2^2,
    \end{split}
\end{equation}
where $\Eb{Y^2}=1$ by the properties of the Rademacher distribution and $\Eb{Yf(\pi_{\{1\}}(\x))} = 0$, because we set $\operatorname{Cov}(Y, \pi_{\{1\}}(X_1)) = 0$ and $\operatorname{Cov}(\pi_{\{1\}}(X_1), X_2) = 0$.
Moreover
\begin{equation}\label{eq:appendix-permutation-importance-x1-part1}
    \begin{split}
         \Eb{f(\pi_{\{1\}}(\x))^2} 
         &= w_1^2\Eb{\pi_{\{1\}}(x_1)} 
         - 2 w_1 w_2 \Eb{\pi_{\{1\}}(x_1) x_2}
         + w_2^2 \Eb{x_2^2} \\
         &= w_1^2(s_1^2 + 1) + w_2^2 s_2^2.
    \end{split}
\end{equation}
Analogously for $\pi_{\{2\}}(\x)$ we obtain
\begin{equation}\label{eq:appendix-permutation-importance-x2}
    \begin{split}
         \Eb{(Y-f(\pi_{\{2\}}(\x)))^2} 
         &= 1 - 2 w_1 + w_1^2(s_1^2 + 1) + w_2^2 s_2^2.
    \end{split}
\end{equation}
\begin{figure}
    \centering
    \subfloat[Permutation Importance for $s_2^2=0.1$]{\label{fig:permutation-s2-01}    \includegraphics[scale=0.4]{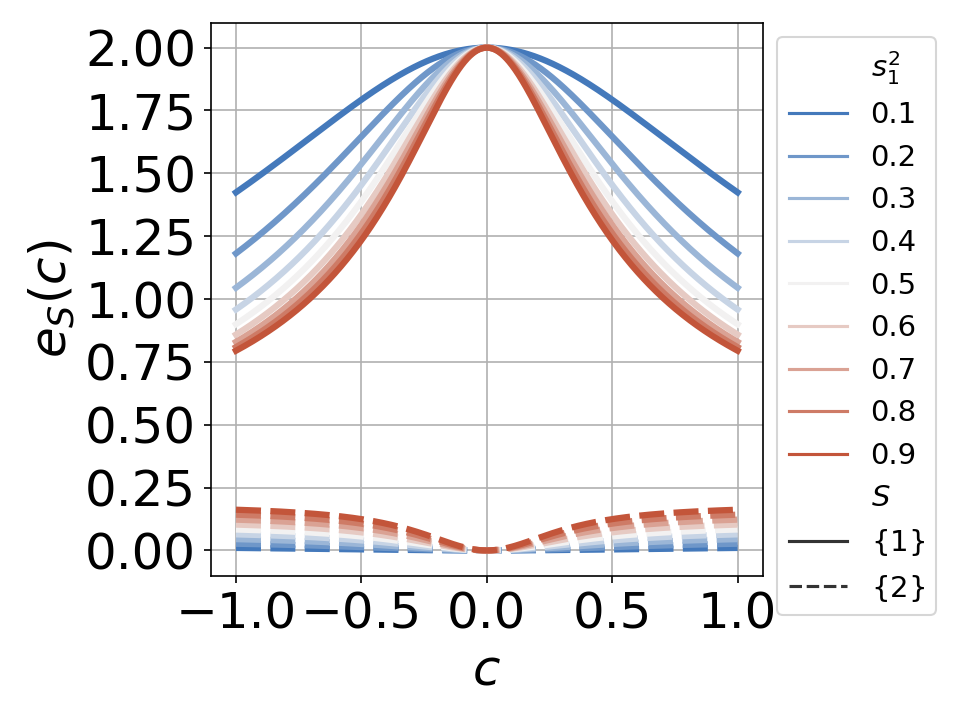}}
\hspace{0.05\columnwidth}
    \subfloat[Permutation Importance for $s_2^2=0.9$]{\label{fig:permutation-s2-09}    \includegraphics[scale=0.4]{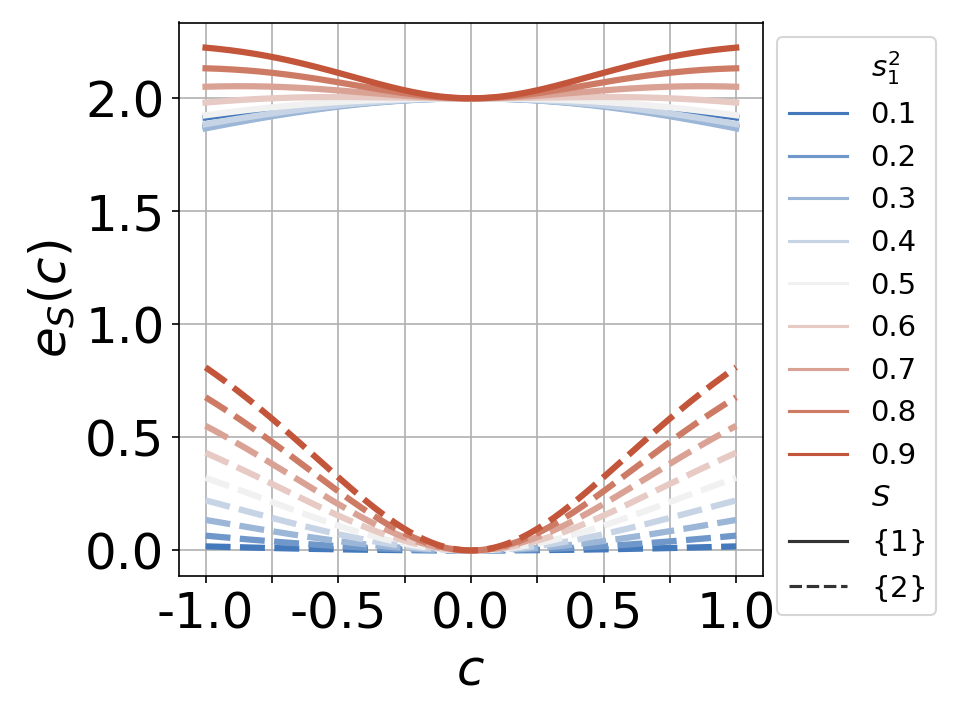}}
       \caption{}
    \label{fig:appendix-permutation}
\end{figure}
%
\section{Conditional expectations}\label{app:cond-expectations}
In order to compute the M-plots $\Eb{w_1x_1 + w_2 x_2| X_{S} = x_{S}}$, we will first consider the conditional expectations $\Eb{X_1 | X_2 = x_2}$ and $\Eb{X_2 | X_1 = x_1}$.
Starting with $\Eb{X_1 | X_2 = x_2}$, by the law of total expectation, we can write
\begin{equation}\label{eq:appendix-conditional-x1-x2}
\begin{split}
        \Eb{X_1 | X_2 = x_2} 
        &= \Eb{X_1 \mid X_2 = x_2, Y =1} \Prob(Y=1\mid X_2 = x_2) \\
        &+ \Eb{X_1 \mid X_2 = x_2, Y =-1} \Prob(Y=-1\mid X_2 = x_2). \\
\end{split}
\end{equation}
Since $\Prob(X_1, X_2 \mid Y = 1) \sim \mathcal{N}((1, 0)^{\top}, \Sigma)$ and $\Prob(X_1, X_2 \mid Y = -1) \sim \mathcal{N}((-1, 0)^{\top}, \Sigma)$, we can straightfowardly compute the conditional expectations
\begin{equation}\label{eq:appendix-conditional-x1-x2-y1-y-1}
\begin{split}
        \Eb{X_1 \mid X_2 = x_2, Y =1} &= \frac{c s_1}{s_2} x_2 \\
        \Eb{X_1 \mid X_2 = x_2, Y =-1} &= \frac{c s_1}{s_2} x_2 .
\end{split}
\end{equation}
And by Bayes' theorem and using the notation of the joint density $p$ (see \eqref{eq:joint-density-x1x2}), we can write
\begin{equation}\label{eq:appendix-conditional-prob-y1-x1}
\begin{split}
        \Prob(Y=1 \mid X_2=x_2) = \frac{p_{1, X_2}(x_2) \Prob(Y=1)}{p_{1, X_2}(x_2)} = \frac{1}{2},
\end{split}
\end{equation}
with marginal density $p_{1, X_2}$ of the marginal distribution of $p$ in $X_2$, i.e.
$p_{1, X_2}(x_2) = \varphi(x_2/s_1)$ with standard normal density $\varphi$. 
In accordance with \eqref{eq:appendix-conditional-prob-y1-x1} we obtain $\Prob(Y=-1 \mid X_2=x_2)=1/2$. Combining the results \eqref{eq:appendix-conditional-x1-x2}, \eqref{eq:appendix-conditional-x1-x2-y1-y-1} and \eqref{eq:appendix-conditional-prob-y1-x1} we yield 
\begin{equation}\label{eq:appendix-conditional-x1-x2-result}
\begin{split}
        \Eb{X_1 | X_2 = x_2} =  \frac{c s_1}{s_2} x_2.
\end{split}
\end{equation}
%
%
%
Again, by the law of total expectation, for $\Eb{X_1 | X_2 = x_2}$, we can compute the conditional expectations $\Eb{X_2 \mid X_1 = x_1, Y =1}$ and $\Eb{X_2 \mid X_1 = x_1, Y =-1}$ in a straightforward manner by the argument used for \eqref{eq:appendix-conditional-x1-x2-y1-y-1}
\begin{equation}\label{eq:appendix-conditional-x2-x1-y1-y-1}
\begin{split}
        \Eb{X_2 \mid X_1 = x_1, Y =1} &= \frac{c s_2}{s_1} (x_1 - 1) \\
        \Eb{X_2 \mid X_1 = x_1, Y =-1} &= \frac{c s_2}{s_1} (x_2 + 1) .
\end{split}
\end{equation}
Furthermore, by Bayes' theorem, we know
\begin{equation}\label{eq:appendix-conditional-prob-y1-x2}
\begin{split}
        \Prob(Y=1 \mid X_1=x_1) 
        &= \frac{p_{1, X_1}(x_1) \Prob(Y=1)}{\frac{1}{2}p_{1, X_1}(x_1)+\frac{1}{2}p_{2, X_1}(x_1)} \\
        &= \vartheta(2x_1/s_1^2),
\end{split}
\end{equation}
where $p_{1, X_1}$ and $p_{2, X_1}$ are the marginal densities of the marginal distribution of $p$ in $X_1$,
namely $p_{1, X_1}(x_1) = \varphi((x_1-1)/s_1)$ and $p_{1, X_1}(x_1) = \varphi((x_1+1)/s_1)$, and sigmoid function $\vartheta:\R \to \R$.
Similarly, $\Prob(Y=-1 \mid X_1=x_1) = 1 - \vartheta(2x_1/s_1^2)$.
The combination of \eqref{eq:appendix-conditional-x2-x1-y1-y-1} and \eqref{eq:appendix-conditional-prob-y1-x2} amounts to \eqref{eq:m-plot-functions}.

\section{Shapley values}\label{app:shapley-values}
With $d=2$ and set of feature indices $[d]$, we consider the Shapley values
\begin{equation}
    \label{eq:app-shapley-value}
    e_{\{j\}} = \sum_{S \subseteq [d] \setminus \{j\}}
    \gamma_{d}(S) \left[ v(S \cup \{j\}) - v(S) \right].
\end{equation}
We define the value function $v$ via a set function $v(S)\coloneqq g_S(\x_S)$, $g_S: \R^{|S|} \to \R$.
For feature sets with two features, the Shapley values are given by
\begin{equation}
    \begin{split}
        e_{\{1\}} 
        = \frac{1}{2}\big( g_{\emptyset \cup \{1\}} - g_{\emptyset}\big)
        + \frac{1}{2}\big( g_{\{1, 2\}} - g_{\{2\}}\big) \\
        e_{\{2\}} 
        = \frac{1}{2}\big( g_{\emptyset \cup \{2\}} - g_{\emptyset}\big)
        + \frac{1}{2}\big( g_{\{1, 2\}} - g_{\{1\}}\big),
    \end{split}
\end{equation}
with their corresponding weights
\begin{equation}
    \begin{split}
            \gamma_{2}(\emptyset) = 1/2, \quad
            \gamma_{2}(\{1\}) = 1/2, \quad \gamma_{2}(\{2\}) = 1/2.
    \end{split}
\end{equation}
In the considered scenarios we use different set functions depending on the particular XAI approach, but set $g_{\emptyset} = 0$.
Now, we state the value functions we used to compute the corresponding Shapley values for each feature.
\paragraph{Coefficient of multiple determination}
For Paragraph \ref{par:coef-determination} we employed the value function $g_S(\x_S) \coloneqq  R_S^2 = \sum_{j \in S} w_j r_j$ and for the corresponding subsets $S$ we yield
\begin{equation}
    \begin{split}
            g_{\emptyset}(\x) &= 0 \quad g_{\{1, 2\}}(\x) = w_1 / (s_1^2 + 1)^{1/2} \quad g_{\{2\}}(\x) = 0 \\
            g_{\{1\}}(\x) &= w_1 / (s_1^2 + 1)^{1/2} \\
            g_{\{1\}}(\x) &= \hat{w}_1 / (s_1^2 + 1)^{1/2} \,\, (\text{for the `sub-model'}f_{\{1\}}(x_1)=\hat{w}_1 x_1).      
    \end{split}
\end{equation}
\paragraph{SHAP}
Using the set function $g_S(\x_S) \coloneqq \E_{x_{C}}(f(\x_S, \x_C))$ and computing the Shapley values for a linear model reduces to
\begin{equation}
    \begin{split}
            g_{S \cup \{j\}}(\x_S) - g_{S}(\x_S) = w_j (x_j - \E(x_j)),     
    \end{split}
\end{equation}
therefore, do not dependent on $S$ \citep[cf.][]{strumbeljExplainingPredictionModels2014}.
For the set function  $g_S(\x_S)\coloneqq\Eb{f(x_{S}, x_{C})| X_{S} = x_{S}}$ consider the derivations provided by \citet{aasExplainingIndividualPredictions2020}.


\section{FIRM}\label{app:firm}

To derive the lower bound for $e_{\{1\}}(\x)$ we first observe that for $x_1 > 0$
\begin{equation}
    \begin{split}
    x_1 - h(x_1) &= x_1 - \left[ (x_1 - 1) \vartheta(2x_1/s_1^2) + (x_1 + 1)(1 - \vartheta(2x_1/s_1^2)) \right] \\
    &= 2 \vartheta(2x_1/s_1^2) - 1\\
    &> 0
    \end{split}
\end{equation}
where $\vartheta(x) := ( 1 + \exp(-x))^{-1}$ denotes the sigmoid function as before.
Thus, we may estimate $e_{\{1\}}(\x)$ from below by
\begin{equation}
    \begin{split}
        e_{\{1\}}(\x)^2 &= \var(\Eb{f(\x) \mid X_1} ) \\
        &= \alpha^2 \Eb{(X_1 - c^2 h(X_1))^2} \\
        &\ge \alpha^2 \Eb{(X_1 - h(X_1))^2 \mid X_1 > 1} \Prob(X_1 > 1) \\
        &= \alpha^2 \Eb{ (2 \vartheta(2X_1/s_1^2) - 1)^2 \mid X_1 > 1} \Prob(X_1 > 1) \\
        &\ge \alpha^2 (2 \vartheta(2/s_1^2) - 1)^2 \Prob(X_1 > 1 \mid Y = 1) \Prob(Y = 1) \\
        &= \frac{\alpha^2}{4} \left(2 \vartheta(2/s_1^2) - 1 \right)^2.
    \end{split}
\end{equation}
Taking the square root on either side now yields the lower bound 
\begin{equation}
    \begin{split}
        e_{\{1\}}(\x) \ge \frac{\alpha}{2} \left(2 \vartheta(2/s_1^2) - 1 \right).
    \end{split}
\end{equation}

\section{LRP and DTD}\label{app:lrp}

The Deep Taylor decomposition \citep[DTD,][]{montavonExplainingNonLinearClassification2017} as a generalization of layerwise relevance propagation \citep[LRP,][]{bachPixelWiseExplanationsNonLinear2015} summarizes this family of explanation methods via the general propagation rule
\begin{equation}
    \label{eq:appendix-dtd-rule}
    e_{\{j\}}(\x) \coloneqq R_{j}^{l-1} = \frac{w \odot (\x-\x_0)}{w^{\top}\x} R_{j}^{l} \;.
\end{equation}
In applications of DTD, the choice of a suitable root point $\x_0$ is of critical importance. 
Here, \citet{kindermansLearningHowExplain2017} observe that in order to extract the `signal' from the data we have to remove the distractor $\eta$ by choosing a signal estimator $S_{\abf}(\x) = x - \eta$, i.e. we pick the root point  $\x_0 = \eta$ and implicitly estimate the distractor $\hat{\eta} = x - S_{\abf}(x)$.
Furthermore, a good signal estimator should yield high values of the quality measure 
\begin{equation}
    \label{eq:signal-quality-measure}
    \rho(S_{\abf}) \coloneqq 1 - \max_{v} \frac{v^{\top} \operatorname{Cov}(\hat{\eta}, y)}{\left( \sigma_{v^{\top} \hat{\eta}}^{2} \sigma_{y}^{2} \right)^{1/2}} \, ,
\end{equation}
i.e. a signal estimator $S_{\abf}$ is optimal if we have a vanishing correlation between $\hat{\eta}$ and $y$.
With these observations \citet{kindermansLearningHowExplain2017} assume a linear dependency between the signal $S_{\abf}$ and the target $y$ yielding a signal estimator 
\begin{equation}
    \label{eq:linear-signal-estimator}
    S_{\abf}(x) =  \abf w^{\top} \x \,.
\end{equation}
Now, consider
\begin{equation}
\label{eq:linear-signal-estimator-derivation}
    \begin{split}
        \operatorname{Cov}(\hat{\eta}, y) &= 0 \\
        \Leftrightarrow \operatorname{Cov}(\x - S_{\abf}(\x), y) &= 0 \\
        \Leftrightarrow  \operatorname{Cov}(\x, y) &= \operatorname{Cov}(S_{\abf}(\x), y) \\
        \Leftrightarrow  \operatorname{Cov}(\x, y) &= \operatorname{Cov}( \abf w^{\top} \x, y) \, ,
    \end{split}
\end{equation}
and for $\operatorname{Cov}( \abf w^{\top} \x, y) = \abf \operatorname{Cov}( y, y)$ we yield the activation pattern $\abf = \operatorname{Cov}( \x, y) / \sigma_{y}^2$.
Applying the signal estimater $S_{\abf}$ to the propagation rule \eqref{eq:appendix-dtd-rule} means replacing $(\x - \x_0)$ by $(\abf w^{\top}\x)$, and with $ R_{j}^{l} = \mathbb{I}_{f(\x) > 0}$ we yield the explanation
\begin{equation}
    \label{eq:appendix-dtd-explanation}
    e_{\{j\}}(\x) = \left( \frac{w \odot (\abf w^{\top}\x)}{w^{\top}\x} \right)_j = \left( w \odot \abf \right)_j \;,
\end{equation}
which is called the \textit{PatternAttribution} method by \citet{kindermansLearningHowExplain2017}. 
The \textit{PatterNet} method only back-propagates the activation patterns $\abf_j$.

\end{document}